\documentclass[10pt,twocolumn,letterpaper]{article}

\usepackage[pagenumbers]{cvpr} 

\usepackage[dvipsnames]{xcolor}

\usepackage[linesnumbered,ruled,vlined]{algorithm2e}

\definecolor{cvprblue}{rgb}{0.21,0.49,0.74}
\usepackage[pagebackref,breaklinks,colorlinks,citecolor=cvprblue]{hyperref}

\newcommand{\papername}{MVHuman }

\title{MVHuman: Tailoring 2D Diffusion with Multi-view Sampling For Realistic 3D Human Generation}

\author{Suyi Jiang \;\, Haimin Luo \;\, Haoran Jiang \;\, Ziyu Wang \;\, Jingyi Yu \;\, Lan Xu\\
\\
ShanghaiTech University\\
}

\begin{document}
\maketitle
\begin{abstract}
Recent months have witnessed rapid progress in 3D generation based on diffusion models. Most advances require fine-tuning existing 2D Stable Diffsuions into multi-view settings or tedious distilling operations and hence fall short of 3D human generation due to the lack of diverse 3D human datasets. We present an alternative scheme named MVHuman to generate human radiance fields from text guidance, with consistent multi-view images directly sampled from pre-trained Stable Diffsuions without any fine-tuning or distilling. Our core is a multi-view sampling strategy to tailor the denoising processes of the pre-trained network for generating consistent multi-view images. It encompasses view-consistent conditioning, replacing the original noises with ``consistency-guided noises'', optimizing latent codes, as well as utilizing cross-view attention layers.
With the multi-view images through the sampling process, we adopt geometry refinement and 3D radiance field generation followed by a subsequent neural blending scheme for free-view rendering. Extensive experiments demonstrate the efficacy of our method, as well as its superiority to state-of-the-art 3D human generation methods.

\end{abstract}
\section{Introduction}
\label{sec:intro}

The 3D creation of us humans with photo-realism serves as the cornerstone for numerous applications like telepresence or immersive experiences in VR/AR. Early attempts~\cite{alexander2009digital} generally require expensive apparatus and immense artistic expertise and hence are limited to celebrities in feature films. Democratizing the accessible use of realistic human avatars to the mass crowd remains unsolved for the vision communities.

\begin{figure}[tbp] 
        \vspace{-1ex}
	\centering 
        \includegraphics[width=1\linewidth]{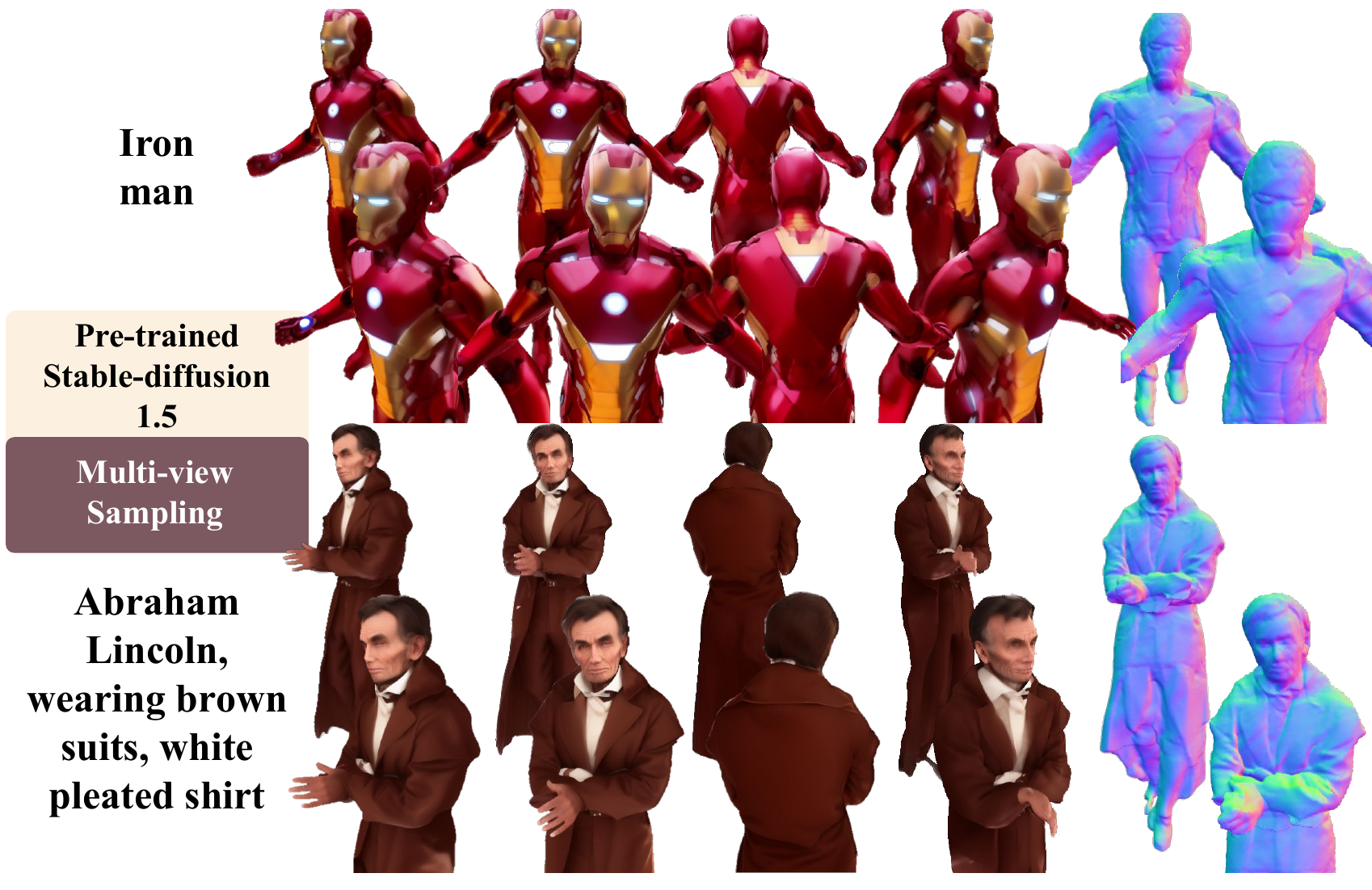}
	\vspace{-20pt} 
	\caption{Given text guidance, MVHuman generates free-view rendering results and fine-grained geometry with the aid of multi-view images sampled from pre-trained 2D diffusion models.} 
	\label{fig:fig_teaser} 
	\vspace{-20pt} 
\end{figure} 

The Diffusion models~\cite{ho2020denoising}, i.e., Stable Diffusion (SD)~\cite{rombach2022high}, have demonstrated high-fidelity and diverse 2D human generation, from simple text prompts. Using ControlNet~\cite{zhang2023adding}, they can even generate hyper-realistic human images under various viewpoints and poses, yielding huge potential for 3D human generation. However, directly training an analogous Diffusion model under 3D representations~\cite{nichol2022point,ssdn_Chen_2023_ICCV} for 3D human generation is infeasible, mainly due to the severe lack of diverse and high-quality human scans. Thus, various 2D-lifting approaches~\cite{Chen_2023_ICCV, lin2023magic3d, poole2022dreamfusion, wang2023score, wang2023prolificdreamer} explore to distill pre-trained 2D diffusion models to optimize certain 3D representations like mesh or NeRF~\cite{nerf}. However, they suffer from inefficient optimization with slow convergence and the multi-faced Janus artifacts due to the lack of 3D awareness. Some recent methods~\cite{cao2023dreamavatar, huang2023dreamwaltz, kolotouros2023dreamhuman, zhang2023avatarverse, liao2023tada} tailor such distillation scheme for 3D human generation. The utilization of human motion and shape priors~\cite{loper2023smpl} alleviates the Janus artifacts, but the inherent inefficiency and the cartoon-like saturated appearances caused by the distillation remain. Instead of directly using the pre-trained SD model~\cite{rombach2022high}, recent methods~\cite{shi2023mvdream,liu2023zero,long2023wonder3d} turn to train multi-view diffusion models, which serves as multi-view priors to significantly improve the subsequent 3D content generation. However, training or fine-tuning such multi-view diffusion models still heavily relies on multi-view image datasets from real-world collection~\cite{yu2023mvimgnet} or CG rendering~\cite{objaverse}. As a result, it's still difficult to extend such multi-view strategies for 3D human generation due to the lack of human datasets.

In this paper, we present \textit{MVHuman} -- a novel scheme to generate human assets from text guidance,
with the aid of consistent multi-view images directly sampled from pre-trained 2D diffusion models (see Fig.~\ref{fig:fig_teaser}). In stark contrast to prior arts, we directly apply an existing 2D latent diffusion model~\cite{rombach2022high} with ControlNet~\cite{zhang2023adding} to obtain such multi-view priors, without tedious fine-tuning or distilling.

The core of our MVHuman is a multi-view sampling process compatible with the 2D diffusion model~\cite{rombach2022high}, which bridges various views by carefully tailoring their input conditions, predicted noises, and the corresponding latent codes.
Specifically, we first apply an off-the-shelf monocular reconstructor~\cite{xiu2023econ} on the generated front-view image to obtain a coarse geometry proxy and project it into various target views to obtain 2D skeletal poses, normal, and depth maps. These 2D attributes serve as the view-consistent conditions for utilizing the ControlNet~\cite{zhang2023adding}.
Secondly, we introduce the concept of ``consistency-guided noise''. Note that in the deterministic sampler~\cite{song2020denoising, lu2022dpm}, with a fixed latent code of a certain sampling step, we can obtain the predicted original signal from the corresponding predicted noise and vice verse. Thus, for sampling processes with different initial random noises, we can blend their predicted original signals at a sampling step, so as to obtain a consistent prediction and transform it back to the corresponding consistency-guided noises of various sampling processes. By replacing the original noises with such consistency-guided ones, various sampling processes will finally recover a consistent signal. We extend such consistency-guided noises into the multi-view setting by treating each view with an individual sampling process. Specifically, for a certain sampling step of a target view, we warp the decoded images from the predicted original signals of its adjacent views to the current view through depth-based warping and then encode these warped images back to the latent space of SD model~\cite{rombach2022high}. We further apply an occlusion-aware blending strategy to obtain the consistent predictions and the subsequent consistency-guided noises, so as to replace the original predicted noises and generate consistent multi-view images.
To further improve the view consistency, we explicitly optimize the latent codes of adjacent views under various sampling steps. We decode them into the image space and warp them to each other through depth-aware warping to calculate their consistency.
Besides, inspired by recent video diffusion methods~\cite{khachatryan2023text2video, ceylan2023pix2video}, we modify the self-attention layer in the pre-trained SD network~\cite{rombach2022high} to concatenate attention features from a reference view. Such a strategy further enhances the appearance similarity of the generated multi-view images.

%


%
Finally, with the above multi-view sampling, we generate the desired human images covering both full-body and upper-body views and subsequently generate radiance fields followed by a blending scheme for free-view rendering. 
We first add fine-grained details into the coarse proxy using the implicit geometric information within the images and then train a radiance field based on multi-plane features~\cite{Cao2022FWD}. We further adapt the neural blending strategy~\cite{suo2021neuralhumanfvv,jiang2022neuralhofusion} into a two-stage coarse-to-fine setting. It blends the initial radiance rendering results with both the full-body and upper-body human images, so as to provide high-fidelity novel view synthesis. As an additional benefit, our MVHuman can seamlessly obtain many functions of the original SD model~\cite{rombach2022high}, such as text-based editing, or loading and upgrading a pre-trained LoRA~\cite{hu2022lora} model into 3D results. 
To summarize, our main contributions include:
\begin{itemize} 
	\setlength\itemsep{0em}
	
	\item We present a novel scheme to generate high-quality human assets, directly using pre-trained 2D diffusion models without fine-tuning or distilling. 
	
	\item We introduce a dedicated multi-view sampling process with consistency-guided noise and latent code optimization, to generate view-consistent images.

        \item We utilize the generated multi-view images to refine geometry and adopt a tiered neural blending scheme on radiance fields to enable free-view rendering.
	
\end{itemize}

\section{Related Work} 
\noindent{\textbf{Text-guided 3D Content Generation.}}
Recent rapid progress in diffusion models within the text-to-image domain~\cite{ho2020denoising, rombach2022high, zhang2023adding} has enhanced research interest in 3D content generation.
Early works~\cite{metzer2022latent, poole2022dreamfusion, 10203874} propose Score Distillation Sampling (SDS) algorithm to lift pre-trained 2D diffusion models to optimize 3D NeRF~\cite{nerf}. 
The following works~\cite{Chen_2023_ICCV, lin2023magic3d} extend SDS to optimize other efficient 3D representations~\cite{Laine2020diffrast, Munkberg_2022_CVPR, shen2021dmtet}, or only generate textures~\cite{Richardson2023TEXTure, chen2023text2tex, lugmayr2022repaint} for existing meshes in pursuit of faster speed and quality. 
However, such SDS-based approaches suffer from oversaturation problems. ProlificDreamer~\cite{wang2023prolificdreamer} addresses such problems via Variational Score Distillation (VSD) but costs much longer optimization.
Another line of works~\cite{liu2023zero, melas2023realfusion, raj2023dreambooth3d, Tang_2023_ICCV, xu2023neurallift, huang2023tech, long2023wonder3d, liu2023one2345, shi2023zero123plus, qian2023magic123, zeng2023ipdreamer} turn to reconstruct 3D content from a single image by distillation. Their challenge evolves into altering the generation distribution by fine-tuning the diffusion network. 
Recent multi-view diffusion models~\cite{shi2023mvdream,tang2023mvdiffusion,liu2023syncdreamer} propose to generate view-consistent 2D images and subsequentially benefit 3D generation significantly, heavily relying on cleaned multi-view dataset.
\begin{figure*}[t] 
    \vspace{-1ex}
	\begin{center} 
            \includegraphics[width=\linewidth]{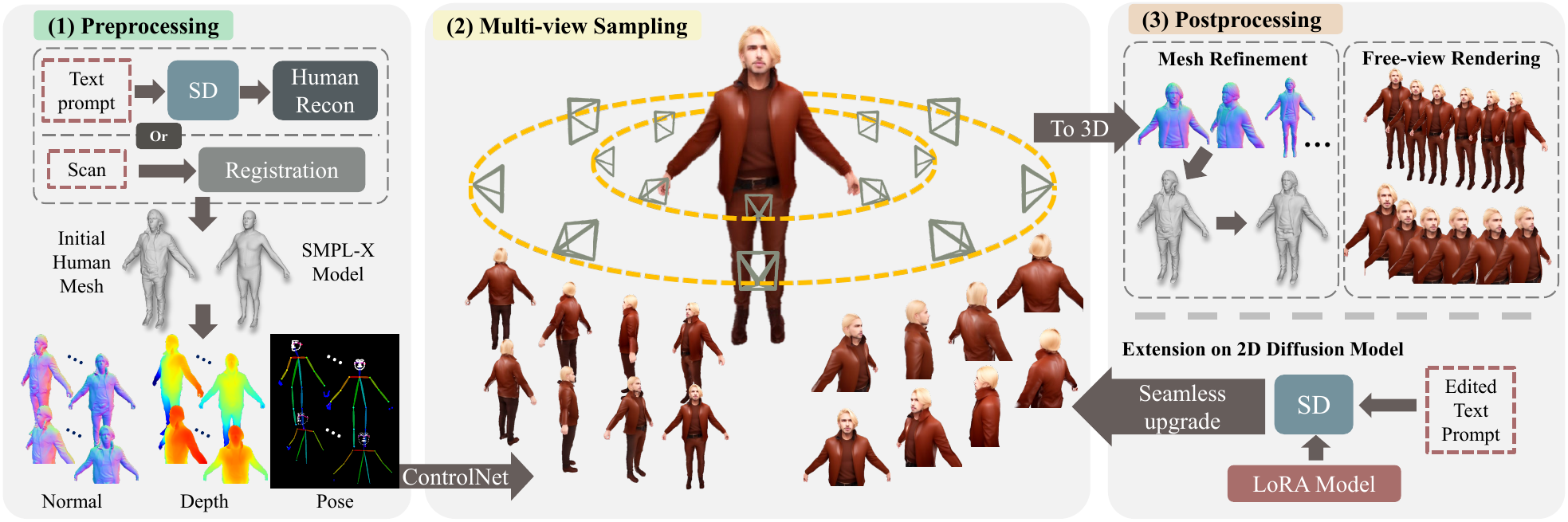} 
	\end{center} 
    \vspace{-20pt}
    \caption{The overview of our MVHuman pipeline. Our method consists of three steps. The first preprocessing step obtains view-consistent conditions from the initial geometry and its aligned SMPL-X (Sec.~\ref{sec:geometry_init}). The second step performs the multi-view sampling process to generate view-consistent images (Sec.~\ref{sec:multiview_sampling}). The final postprocessing step includes geometry refinement and free-view rendering (Sec.~\ref{sec:postprocessing}).} 
	\label{fig:fig_overview} 
	\vspace{-10pt}
\end{figure*}

\noindent{\textbf{3D Human Generation.}}
Early 3D generative models based on textured mesh~\cite{oechsle2019texture, Liao2020CVPR, hong2022avatarclip}, point cloud~\cite{achlioptas2018learning, yang2019pointflow}, and voxels~\cite{nguyen2019hologan, nguyen2020blockgan, zhou2021cips} suffer from limited expressiveness while requiring 3D datasets.
Recent GAN-based works~\cite{hong2022eva3d,jiang2023humangen,zhang2022avatargen,zhang20223d} utilize NeRF-like representations~\cite{EG3D} to achieve directly 3D human generation using 2D images~\cite{fu2022styleganhuman}.
%
%
Recent diffusion-based works~\cite{jiang2023avatarcraft,cao2023dreamavatar,zhang2023avatarverse,huang2023dreamwaltz} combine various 3D representation and human priors~\cite{alldieck2021imghum,jiang2022text2human,SMPL2015,SMPLX2019} with SDS method~\cite{poole2022dreamfusion} and achieve better quality.
AvatarCraft~\cite{jiang2023avatarcraft} leverages NeuS\cite{wang2021neus,long2022sparseneus} and the SMPL model to facilitate the generation of avatars.
DreamHuman~\cite{kolotouros2023dreamhuman} uses imGHUM~\cite{alldieck2021imghum} to constrain a deformable NeRF for human.
Similarly, DreamAvatar~\cite{cao2023dreamavatar}, DreamWaltz~\cite{huang2023dreamwaltz}, and AvatarVerse~\cite{zhang2023avatarverse} utilize SMPL as shape prior.
TADA~\cite{liao2023tada} creates displacement and texture layers based on the SMPL-X model to generate 3D avatars.
These SDS-based methods also inherit limitations that tend to generate oversaturated appearance, which is a severe artifact for human generation.

\noindent{\textbf{3D Human Reconstruction.}}
Many previous works~\cite{PIFU_2019ICCV, PIFuHD, huang2020arch, he2021arch++} achieve monocular human geometry reconstruction using implicit function, but they are unstable for novel human poses. 
ICON~\cite{xiu2022icon} and ECON~\cite{xiu2023econ} take the SMPL to guide explicit normal estimation and achieve a better trade-off between geometry detail and pose stability. 
Other works~\cite{zhao2022humannerf, noguchi2021neural, peng2021neural, wang2022arah, li2022tava} combine NeRF techniques~\cite{mildenhall2020nerf, wang2021ibrnet, chen2022tensorf, meng2021gnerf, 9466273, mueller2022instant, 10.1145/3528223.3530086, Cao2022FWD, kerbl3Dgaussians} with human shape prior to model both human geometry and appearance, while a neural texture blending scheme~\cite{zhao2022humannerf,jiang2022neuralhofusion, 10.1145/3528223.3530086} has been demonstrated to enhance the realisim of appearance. 
In this work, we adopt ECON to provide shape prior for human image generation and produce high-fidelity human assets.

\section{Method}\label{sec:algorithm}

Here, we introduce our text-guided human radiance fields generation scheme, MVHuman, which seeks to directly utilize the 2D generative capability of the existing latent diffusion model~\cite{rombach2022high} with ControlNet~\cite{zhang2023adding} without extra fine-tuning or distilling. 
As illustrated in Fig.~\ref{fig:fig_overview}, the core of our \papername is a novel multi-view sampling process to simultaneously sample multiple view-consistent images with the aid of a coarse human geometry proxy (Sec.~\ref{sec:geometry_init}). 
Specifically, we construct ``consistency-guided noise'' in sampling steps to gradually denoise the individual initial random noises of multiple views into consistent ones (Sec.~\ref{sec:multiview_sampling}). 
With the multi-view sampling above, we carefully generate high-quality human images from multiple view points which enable reconstructing detailed 3D geometry and generating neural radiance fields followed by a neural blending scheme for free-view rendering (Sec.\ref{sec:postprocessing}). 

\subsection{Preprocessing}\label{sec:geometry_init} 

We propose to generate a rough human mesh as a geometry proxy from scratch guided by a text prompt. 
Specifically, we adopt the stable diffusion model to generate a realistic full-body human front-view image and then utilize the off-the-shelf ECON~\cite{xiu2023econ} model to reconstruct full-body mesh and aligned SPML-X~\cite{SMPLX2019} model providing 3D pose.
Then we project them to desired target views to depth, normal maps, and 2D skeletal poses, providing view-consistent conditions for the ControNet~\cite{zhang2023adding} in the following sampling process.
Note that we also support pre-provided geometry, e.g., human scans, by utilizing the registration technique~\cite{bhatnagar2020loopreg} to obtain the corresponding SMPL-X model.

%

%

%
%


\subsection{Multi-view Sampling Process}\label{sec:multiview_sampling} 
Here we describe our multi-view sampling process to generate view-consistent human images for various views, e.g., views evenly placed on circular tracks shown in Fig.~\ref{fig:fig_overview}, with the aid of geometry proxy and controlling conditions. 
The key observation is, for simultaneous sampling processes with different initial random noises, if we blend their predicted original signals and transform it back to a novel noise to replace the original predicted ones at each time step, such sampling processes will finally recover consistent signals (refer to the supplemental materials for more details).
We apply such modified noise to a multi-view setting called "consistency-guided noise" so as to constrain the visible parts from multiple viewpoints to be as consistent as possible. 
Besides, we propose a latent codes optimization scheme and apply feature concatenation to the self-attention blocks in the SD model to further enhance the view consistency of sampled human images. 

\noindent{\textbf{Preliminary.}}
Stable diffusion~\cite{rombach2022high} is a diffusion model that learns a denoising backward process in the latent space of an encoder $\mathcal{E}$ and a decoder $\mathcal{D}$. 
It adopts a UNet-like network $\boldsymbol{\epsilon}$ conditioned on a text-prompt embedding to predict the noise $\epsilon_t$ at each time step $t$ in the denoising process, which should follow a normal distribution of $\mathcal{N}(\textbf{0}, \textbf{I})$. 
A deterministic sampling process such as DDIM~\cite{song2020denoising} is applied to denoise latent $x_t$ to $x_{t-1}$ as:
\begin{equation}
x_{t-1}=\sqrt{\alpha_{t-1}}x_{0\leftarrow t}+ \\ \sqrt{1-\alpha_{t-1}} \epsilon_t, t=T, \ldots, 1,
\label{eqn:eqn_samplestep} 
\end{equation}
where $\left\{\alpha_t\right\}_{t=1}^T$ are constants from DDIM~\cite{song2020denoising} and ${x}_{0\leftarrow t}$ is the predicted original signal at step $t$:
\begin{equation}
{x}_{0\leftarrow t}=\left(\frac{x_t-\sqrt{1-\alpha_t} \epsilon_t}{\sqrt{\alpha_t}}\right).
\label{eqn:eqn_noise2origin} 
\end{equation}
We denote Eqn.~\ref{eqn:eqn_samplestep} as: 
$x_{t-1}=\mathcal{S}(x_t, \epsilon_t)$.
Note that we omit the conditional inputs, e.g., text, depth, for brevity.

\begin{figure*}[t] 
    \vspace{-1ex}
	\begin{center} 
		\includegraphics[width=\linewidth]{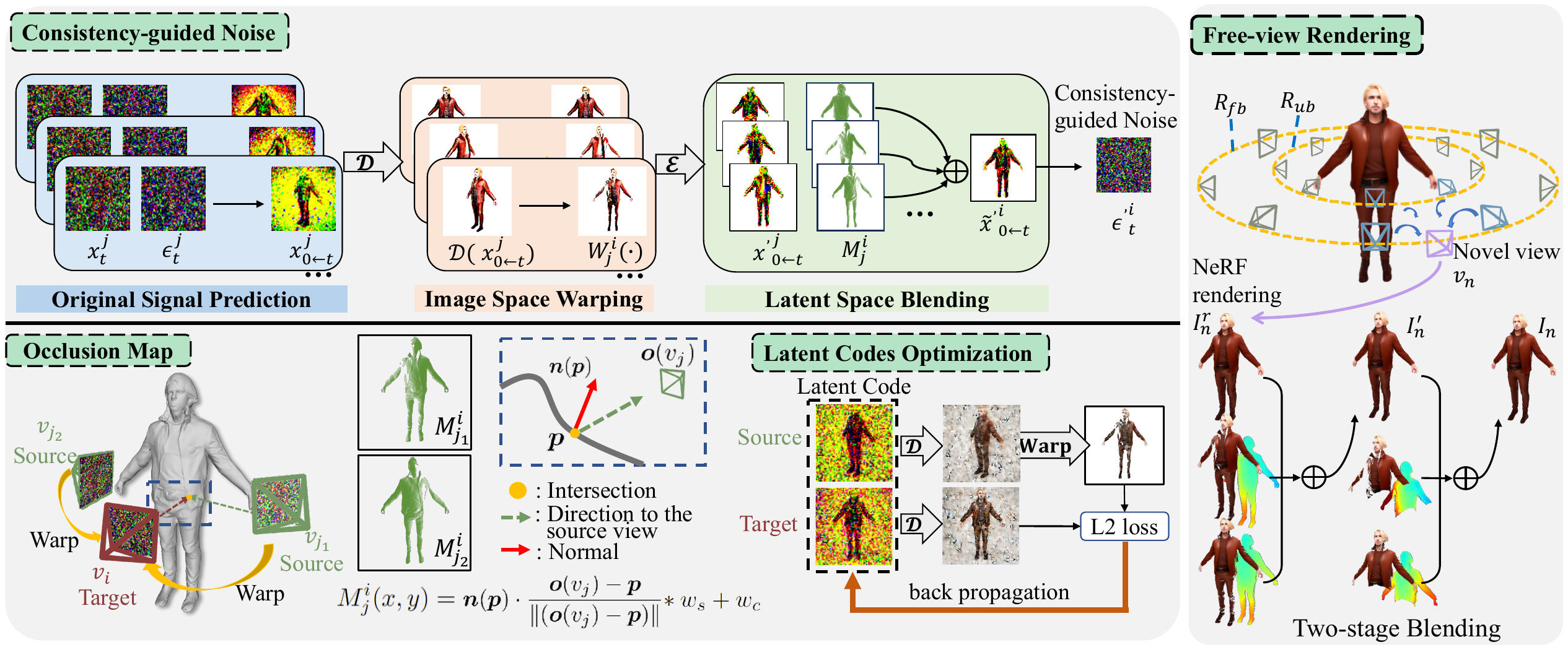} 
	\end{center} 
    \vspace{-20pt}
    \caption{Illustration of our method. The left-top area illustrates the construction of the consistency-guided noise for each view with its adjacent source views at a sampling step. The left-down area illustrates the generation of weighted occlusion maps. The mid-down area illustrates the optimization of latent codes. And the right area illustrates the adopted two-stage blending scheme on human radiance fields for free-view rendering.}
	\label{fig:fig_method} 
	\vspace{-10pt}
\end{figure*}

\noindent{\textbf{Consistency-guided Noise.}}
As shown in Fig.~\ref{fig:fig_method}, we propose to construct consistency-guided noise $\epsilon^{' i}_t$ at sampling steps for each view $v_i$ in total $N$ views thus gradually denoise the latent codes $x^i_t$ to view-consistent ones as: 
$x^i_{t-1}=\mathcal{S}(x^i_t, \epsilon^{' i}_t).$
To this end, we first conduct a warping-based blending scheme to fuse the predicted original signals 
${x}_{0\leftarrow t}^{j_1}, {x}_{0\leftarrow t}^{j_2}, \ldots$ 
of adjacent source views $v_{j_1}, v_{j_2}, \ldots$ to the target view $v_i$ using the initial human mesh.  
Specifically, we warp the decoded images from source views to target view and then encode them back to latent space rather than warping in latent space to avoid misalignment:
\begin{equation}
    {x}^{' j}_{0\leftarrow t} = \mathcal{E}(W^i_{j}(\mathcal{D}({x}^{j}_{0\leftarrow t})))
    \label{eqn:eqn_space_transfer} 
\end{equation}
where $W^i_{j}$ denotes the warping operation from $v_j$ to $v_i$.
Since the warping operation will not modify the pixel values, we can assume that the transformed original signal ${x}^{' j}_{0\leftarrow t}$ still follows a normal distribution of $\mathcal{N}(\mu_j, \sigma^2)$ (refer to the supplemental materials for more details).
Then we generate weighted occlusion maps $M_j^i$ from $v_j$ to $v_i$ according to the visibility shown in Fig.~\ref{fig:fig_method}:
\begin{equation}
M_j^i(x,y) = \left\{
\begin{aligned}
& 0,  \text{if $\boldsymbol{p}$ is invisible to $v_j$} \\
& (\boldsymbol{n}(\boldsymbol{p})\cdot \frac{\boldsymbol{o}(v_j)-\boldsymbol{p}}{\|(\boldsymbol{o}(v_j)-\boldsymbol{p})\|}) * w_s + w_c, \text{else}  
\end{aligned}
\right.
\label{eqn:eqn_transfer} 
\end{equation}
where $\boldsymbol{p}$ is the intersection point of the ray casting from the pixel $(x, y)$ of $v_i$ and the human mesh, $\boldsymbol{n}(\boldsymbol{p})$ is the corresponding normal and $\boldsymbol{o}(v_j)$ is the camera position of $v_j$. $w_s$ is a scaling factor and $w_c$ is a constant to avoid numerical instability.
Now we can get the weighted average $\bar{{x}}^{' i}_{0\leftarrow t}$ of these processed original signals for target view $v_i$:
\begin{equation}
\begin{aligned}
\bar{{x}}^{' i}_{0\leftarrow t} &=\frac{M^i_i}{M_{sum}}  {x}^{' i}_{0\leftarrow t} + \frac{M^i_{j_1}}{M_{sum}}  {x}^{' j_1}_{0\leftarrow t} + \ldots, \\
M_{sum} &= M^i_i + M^i_{j_1} + \ldots .
\end{aligned}
\end{equation}

Here the $\bar{{x}}^{' i}_{0\leftarrow t}$ should follow the distribution of 
$\mathcal{N}(\frac{M^i_i}{M_{sum}}\mu_i + \frac{M^i_{j_1}}{M_{sum}}\mu_{j_1} + \ldots, \frac{{M_i^i}^2 + {M_i^{j_1}}^2 + \ldots}{M_{sum}^2}\sigma^2)$.
Ideally we should scale the noise part of ${x}^{' i}_{0\leftarrow t}, {x}^{' j_1}_{0\leftarrow t}, \ldots$ with factor $\frac{M_{sum}}{\sqrt{{M_i^i}^2 + {M_i^{j_1}}^2 + \ldots}}$ to maintain the variance unchanged as $\sigma^2$. However, $\mu_i, \mu_{j_1}, \ldots$ and $\sigma^2$ cannot be easily measured because the decode-then-encode operation cannot guarantee the fidelity. Thus we empirically use the following formula to blend a new weighted average $\Tilde{{x}}^{' i}_{0\leftarrow t}$ shown in Fig.~\ref{fig:fig_method}:
\begin{equation}
\Tilde{{x}}^{' i}_{0\leftarrow t} = \sum_{k=i,j_1,\ldots} \frac{M^i_k}{M_{sum}}  (E{x}^{' k}_{0\leftarrow t} + (1-E)\bar{{x}}^{' k}_{0\leftarrow t}) ,
\label{eqn:eqn_empirical}
\end{equation}
where $E=\frac{M_{sum}}{\sqrt{{M_i^i}^2 + {M_i^{j_1}}^2 + \ldots}}$.

Finally we can get the consistency-guided noise using Eqn.\ref{eqn:eqn_noise2origin}: $\epsilon_{t}^{' i}=\frac{x^i_t - \sqrt{\alpha_t}\Tilde{{x}}^{' i}_{0\leftarrow t}}{\sqrt{1-\alpha_t}}$ which can be used to sample the latent of step $t-1$ for view $v_i$: $x^i_{t-1}=\mathcal{S}(x^i_t, \epsilon_{t}^{' i})$.

\noindent{\textbf{Optimization of Latent Codes.}}
We further optimize the latent codes to enforce view consistency across different views. 
As shown in Fig.~\ref{fig:fig_method}, for a target view $v_i$ and a source view $v_j$, we decode the latent code $x^j_t$ of $v_j$ to image space and then warp the image to $v_i$, and then apply L2 loss on the covered pixels in image space.
We also conduct the same operation in reverse to formulate the following loss to optimize both $x^i_t$ and $x^j_t$:
\begin{equation}
\mathcal{L}_{latent} = \|\mathcal{D}(x^i_t) - W^i_j(\mathcal{D}(x^j_t))\|_2^2 + \|\mathcal{D}(x^j_t) - W^j_i(\mathcal{D}(x^i_t))\|_2^2.
\label{eqn:optim}
\end{equation}
\noindent{\textbf{Self-attention Layer Modification.}}
To further enhance the view similarity between adjacent views, we also import consistency prior to views following recent video diffusion methods~\cite{khachatryan2023text2video, ceylan2023pix2video} by modifying the self-attention blocks in the SD model.
More specifically, for a view $v_i$, the input features to the self-attention block are query $Q^i$, key $K^i$, and value $V^i$. We extend the block to concatenate value and key from a predefined reference view, e.g., facing view, with the original ones:
\begin{equation}
\begin{aligned}
\text{Ex-Attn}\left(Q^i, [K^{ref}, K^i], [V^{ref}, V^i]\right)= & \\
\text{Softmax}\left(\frac{Q^i\left([K^{ref}, K^i]\right)^T}{\sqrt{c}}\right)[V^{ref}, V^i]. &
\end{aligned}
\end{equation}

\noindent{\textbf{Human Images Generation.}}
As depicted in Fig.\ref{fig:fig_method}, we configure two concentric circular tracks of viewpoints, each uniformly distributed with $N$ cameras (we set $N=8$).
The first track of views looks at the full body, denoted as $R_{fb}=\{v_i| i=1,\ldots,N\}$ and the second focuses on the upper body only, denoted as $R_{ub}=\{v_i| i=N+1,\ldots,2N\}$. 
For each view $v_i$ on $R_{ub}$ as the target view, its neighboring views on the same track within a range of $60^\circ$ clockwise and counterclockwise are used as its source views.
For each view $v_i$ on $R_{fb}$ as the target view, in addition to its neighboring views on the same track, we set the view $v_j$ ($j=i+N$) from $R_{ub}$ with the same azimuth angle as an additional source view. 
We replace the $\Tilde{{x}}^{' i}_{0\leftarrow t}$ with $W^i_j({x}_{0\leftarrow t}^{' j})$ in the region where $M_j^i > 0$ to maintain the distinguishable details, e.g., facial details, from the closeup views.
Throughout the multi-view sampling process, we alternate between leveraging predicted noise and consistency-guided noise. To be precise, we sample with $\epsilon^i_t$ when $t$ is odd and $\epsilon^{' i}_t$ when $t$ is even. We find it helps enhance the fidelity of the generated details.

For the latent optimization, we initially compute the loss following Eqn.~\ref{eqn:optim} for each pair of adjacent views on the same track $R_{fb}$ or $R_{ub}$. 
To improve generated details, we lock the latent codes from the front-view and back-view and further lock those from the side-views upon completion of $20\%$ of the sampling process.
Subsequently, we optimize between the two tracks, for each view $v_i$ on $R_{fb}$, we compute the loss with the locked latent code from $v_{i+N}$ on $R_{ub}$.
More details are available in the supplementary material.
\subsection{Postprocessing for Human Assets}\label{sec:postprocessing}
%
%

\noindent{\textbf{Geometry Optimization.}}
In order to facilitate the generation of 3D human assets, we leverage the implicit geometric information within the human images to refine the initial mesh. 
For the obtained desired high-quality human images as $\{I_i|i=1,\ldots,2N\}$ from the multi-view sampling process, we further utilize the human normal predictor from ECON to predict corresponding normals, denoted as $\{\mathcal{N}^p_i|i=1,\ldots,2N\}$.
We then utilize a differentiable rasterizer~\cite{worchel:2022:nds, Laine2020diffrast} to render the normal map of the initial mesh to each view, denoted as $\{\mathcal{N}^m_i|i=1,\ldots,2N\}$.
Subsequently, we optimize the position of vertices to minimize the difference between the predicted normal maps and rendered ones~\cite{kim2023chupa}, however, the predicted normals might not conform strictly to the value range typical of an authentic normal map. Thus, we approximate gradients of normal maps using the Sobel operator $G$ and conduct L2 loss in gradient space as:
\begin{equation}    
\mathcal{L}_{normal} = \sum^{2N}_{i=1}\|G(\mathcal{N}^m_i) - G(\mathcal{N}^p_i)\|_2^2.
\end{equation}

\begin{figure*}[t] 
    \vspace{-1ex}
	\begin{center} 
		\includegraphics[width=\linewidth]{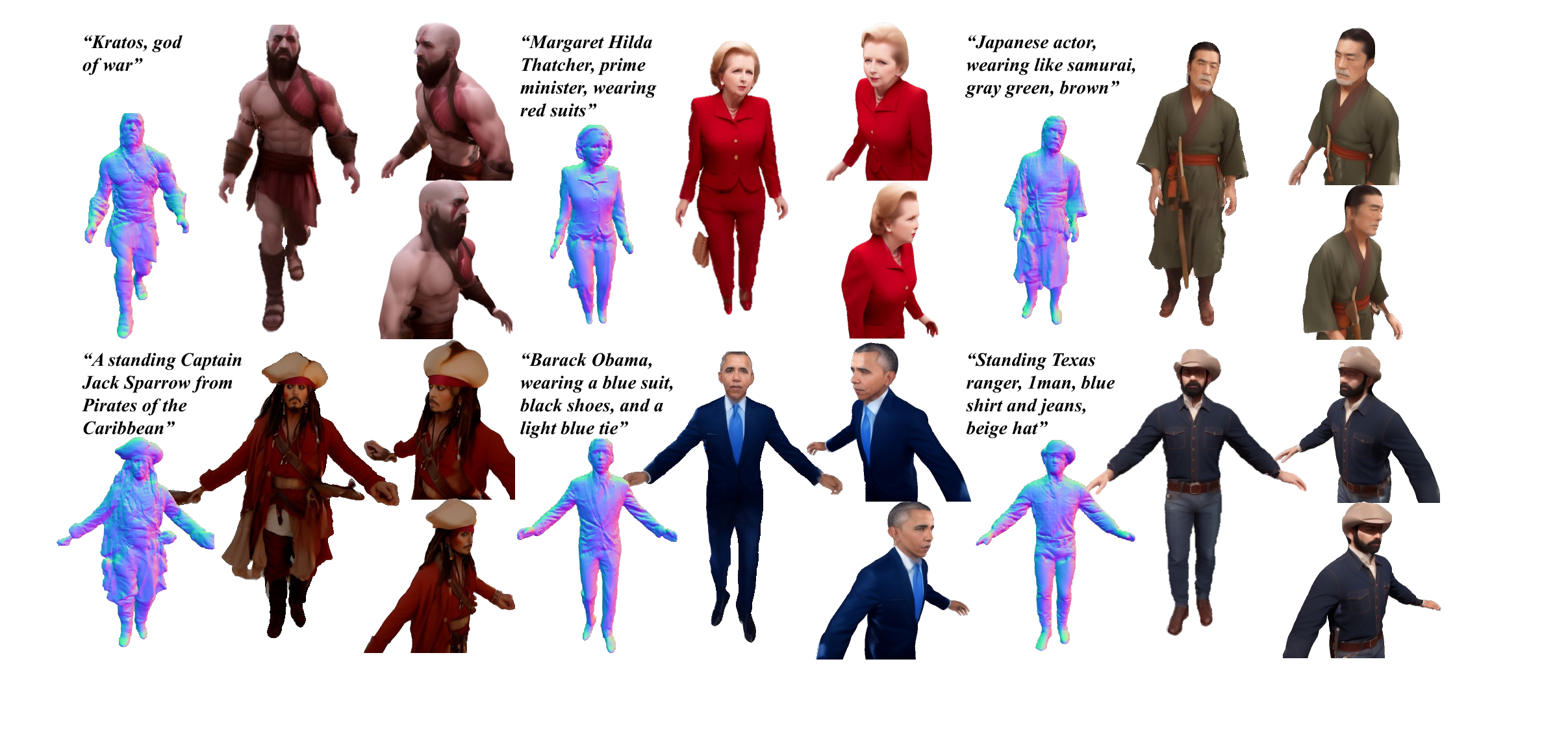} 
	\end{center} 
    \vspace{-20pt}
    \caption{The texture and geometry results of MVHuman, including characters from games/movies, celebrities, and customized humans.} 
	\label{fig:fig_gallery} 
	\vspace{-10pt}
\end{figure*} 

\noindent{\textbf{Free-view Rendering.}}
With multi-view images, we are already able to train an efficient neural radiance field utilizing a multi-plane representation~\cite{Cao2022FWD}. However, the rendering quality has yet to reach the level of fidelity exhibited by the images themselves. 
Thus we further implement a neural blending strategy characterized by a two-stage, coarse-to-fine approach that initially integrates the full-body views followed by the inclusion of the upper-body views.
Following ~\cite{suo2021neuralhumanfvv}, we train a UNet-like network $\Theta$ that takes warped RGB and depth disparity maps as input and outputs two blending weights: $[W_1, W_2] = \Theta(\hat{I}_1, \hat{I}_2, O_1, O_2)$.
As shown in Fig.~\ref{fig:fig_method}, given a novel view $v_n$, we first obtain the NeRF rendering result $I^{r}_n$ at $v_n$ and the rendered depth map $D_n$ from mesh.
In the first stage, we find two nearest views to $v_n$ from full body views $R_{fb}$, denoted as $v_{j_1}, v_{j_2}$. We then warp their images to the novel view and obtain the depth disparity maps from the warped depth and $D_n$ as inputs to
$\Theta$: $[W_{j_1}, W_{j_2}] = \Theta(\hat{I}_{j_1}, \hat{I}_{j_2}, O_{j_1}, O_{j_2})$. 
The blended result of the first stage is denoted as $I^{'}_n$.
In the second stage, we further find two nearest views to $v_n$ from upper body views $R_{ub}$, denoted as $v_{j_3}, v_{j_4}$. Employing a similar procedure, we calculate their blending weights $W_{j_3}, W_{j_4}$. To this end, we get the final blended result $I_n$ as:
\begin{equation}
\begin{aligned}
I^{'}_n= & W_{j_1}\odot \hat{I}_{j_1} + W_{j_2}\odot \hat{I}_{j_2} + (1-W_{j_1}-W_{j_2})\odot I^r_n\\
I_n = &W_{j_3}\odot \hat{I}_{j_3} + W_{j_4}\odot \hat{I}_{j_4} + (1-W_{j_3}-W_{j_4})\odot I^{'}_n
\end{aligned}
\end{equation}

\section{Experimental Results} 
In this section, we first demonstrate the capability of our MVHuman, and then compare our method against the state-of-the-art 3D human generation methods by conducting a user study.
We further evaluate each component of our method on view consistency and quality with quantitative and qualitative results.

\begin{figure*}[t] 
    \vspace{-1ex}
	\begin{center} 
		\includegraphics[width=\linewidth]{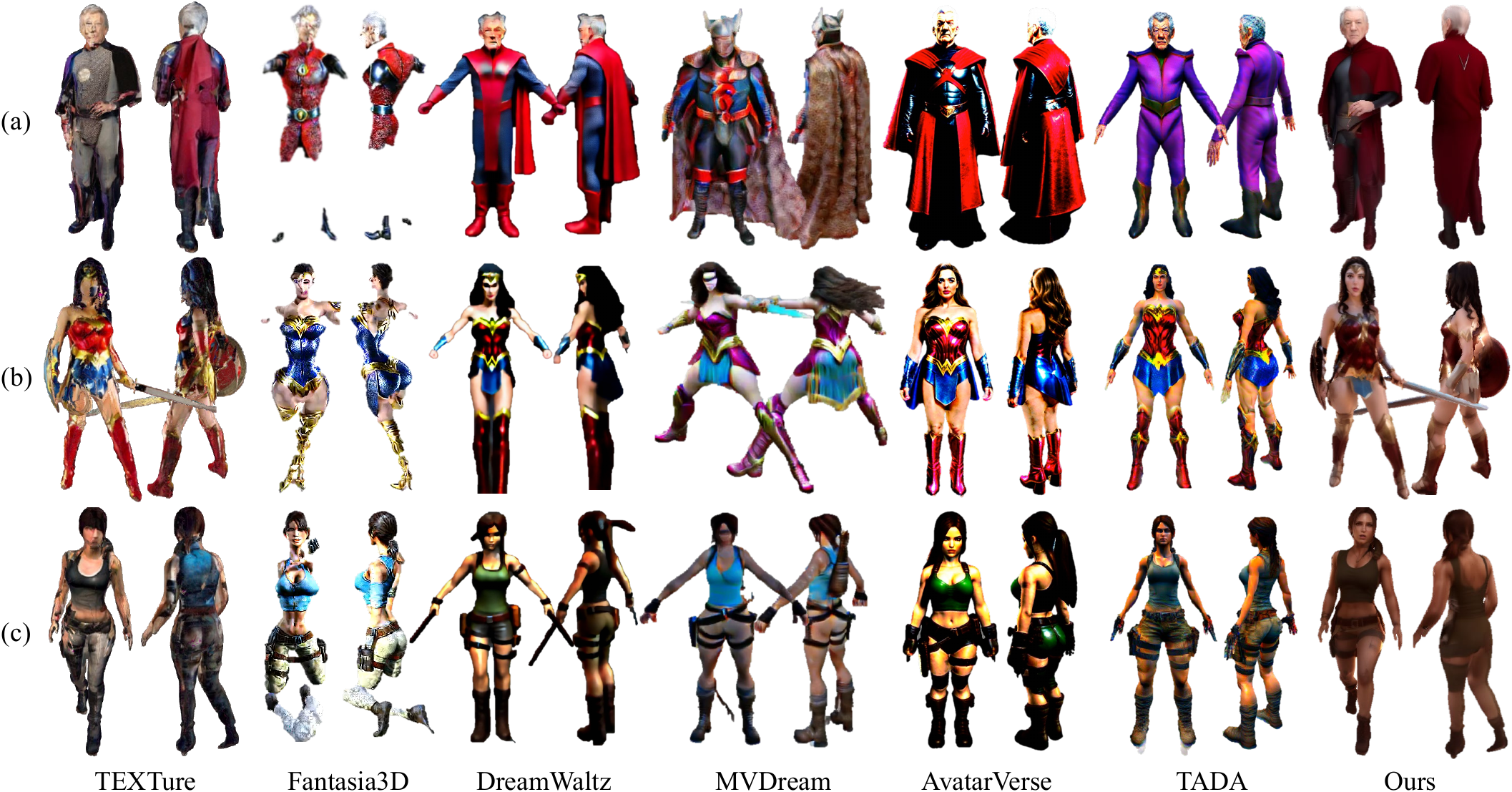} 
	\end{center} 
    \vspace{-20pt}
    \caption{Qualitative comparison between TEXTure, Fantasia3D, DreamWaltz, MVDream, AvatarVerse, TADA and our MVHuman. Our method balances geometric quality and appearance details while avoiding oversaturation and Janus Failure. The prompts from (a) to (c) are \textit{``Ian McKellen, Magneto''}, \textit{``Gal Gadot, Wonder Woman''}, \textit{``Lara Croft, Tomb Raider''}. (* Our second is a case using scan mesh.)}
	\label{fig:fig_qualitative_comparison} 
	\vspace{-10pt}
\end{figure*} 

\subsection{Comparison} 
We compare our method with state-of-the-art text-guided 3D generation methods, TEXTure~\cite{Richardson2023TEXTure}, Fantasia3D~\cite{Chen_2023_ICCV}, DreamWaltz~\cite{huang2023dreamwaltz}, MVDream~\cite{shi2023mvdream}, AvartarVerse~\cite{zhang2023avatarverse} and TADA~\cite{liao2023tada}. 
As illustrated in Fig.~\ref{fig:fig_qualitative_comparison}, 
TEXTure generates texture with ghosting and suffers from Janus Failure. 
Fantasia3D fails to generate complete human body geometry. 
MVDream generates reasonable geometries but its results lack detail, especially in the head region.
DreamWaltz generates blurry texture with unclear edges.
TADA takes more than 3 hours for each prompt, and it fails to generate geometries with correct proportions.
The authors of AvatarVerse provide us with the cases for qualitative comparison.
Their method generates detailed and realistic geometry, but the color of its texture is oversaturated.

\begin{figure}[t] 
    \vspace{-2ex}
	\begin{center} 
		\includegraphics[width=0.93\linewidth]{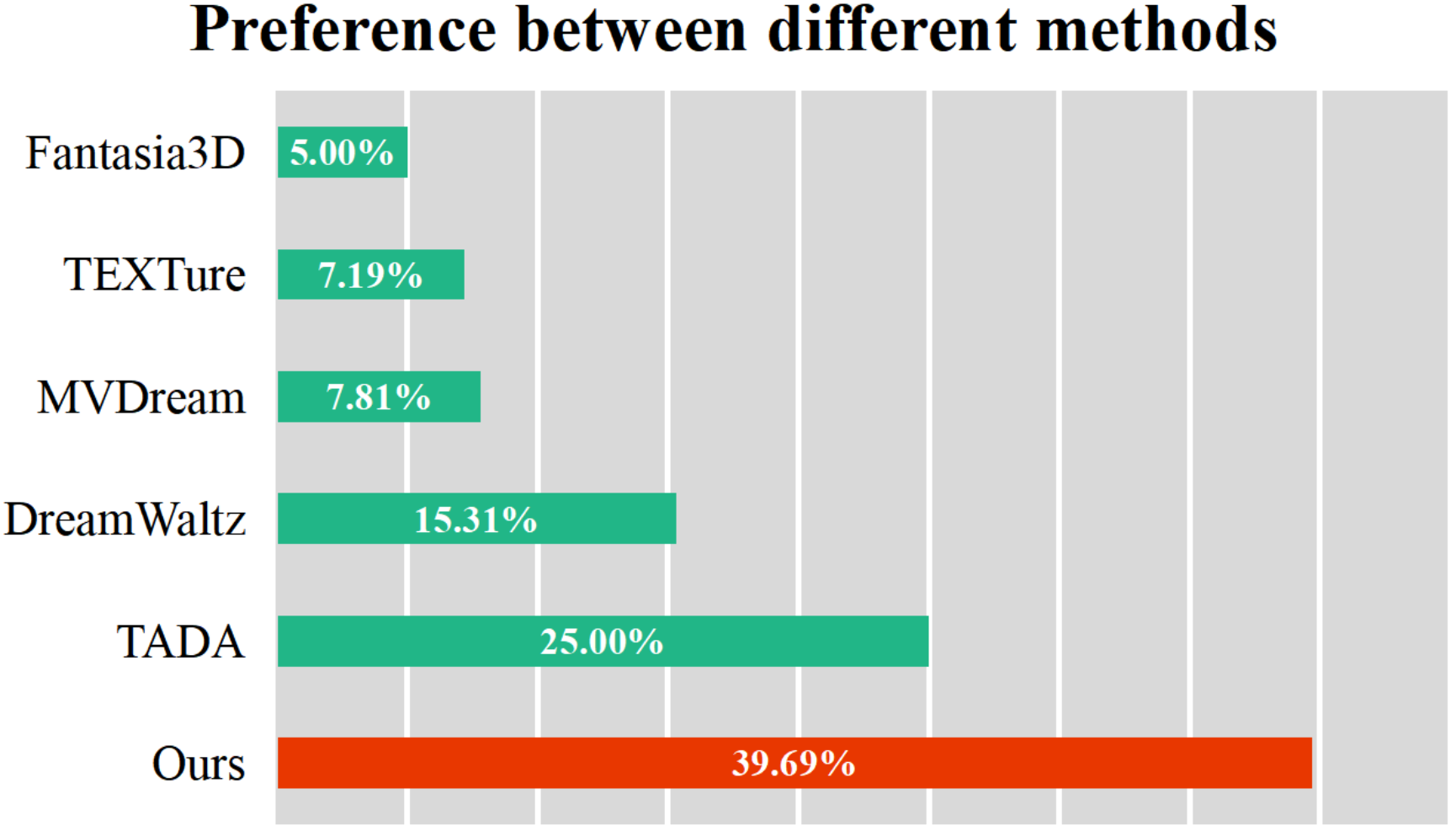} 
	\end{center} 
    \vspace{-20pt}
    \caption{Result of user study over Fantasia3D, TEXTure, MVDream, DreamWaltz, TADA, and our method.}
	\label{fig:fig_quatitative_comparison} 
	\vspace{-10pt}
\end{figure}

We also conduct a comprehensive user study to evaluate the performance of our generated human assets in terms of whether they match the given prompts and their overall quality. 30 prompts describing well-known people and characters are designed and used to generate and render 360-degree videos with full-body and upper-body views. We randomly select 10 samples for each user and ask them to comprehensively consider and select the best results for each sample from four aspects: conforming to the prompt text, geometric and texture quality, face details, and body proportions. As illustrated in Fig.~\ref{fig:fig_quatitative_comparison}, 26 volunteers majoring in computer vision completed the survey, and our method shows significant advantages in the result. Please refer to the supplemental materials for more comparisons.

\subsection{Ablation Study} 
\begin{figure}[t] \label{sec:abla_add} 
    \vspace{-1ex}
	\begin{center} 
		\includegraphics[width=0.95\linewidth]{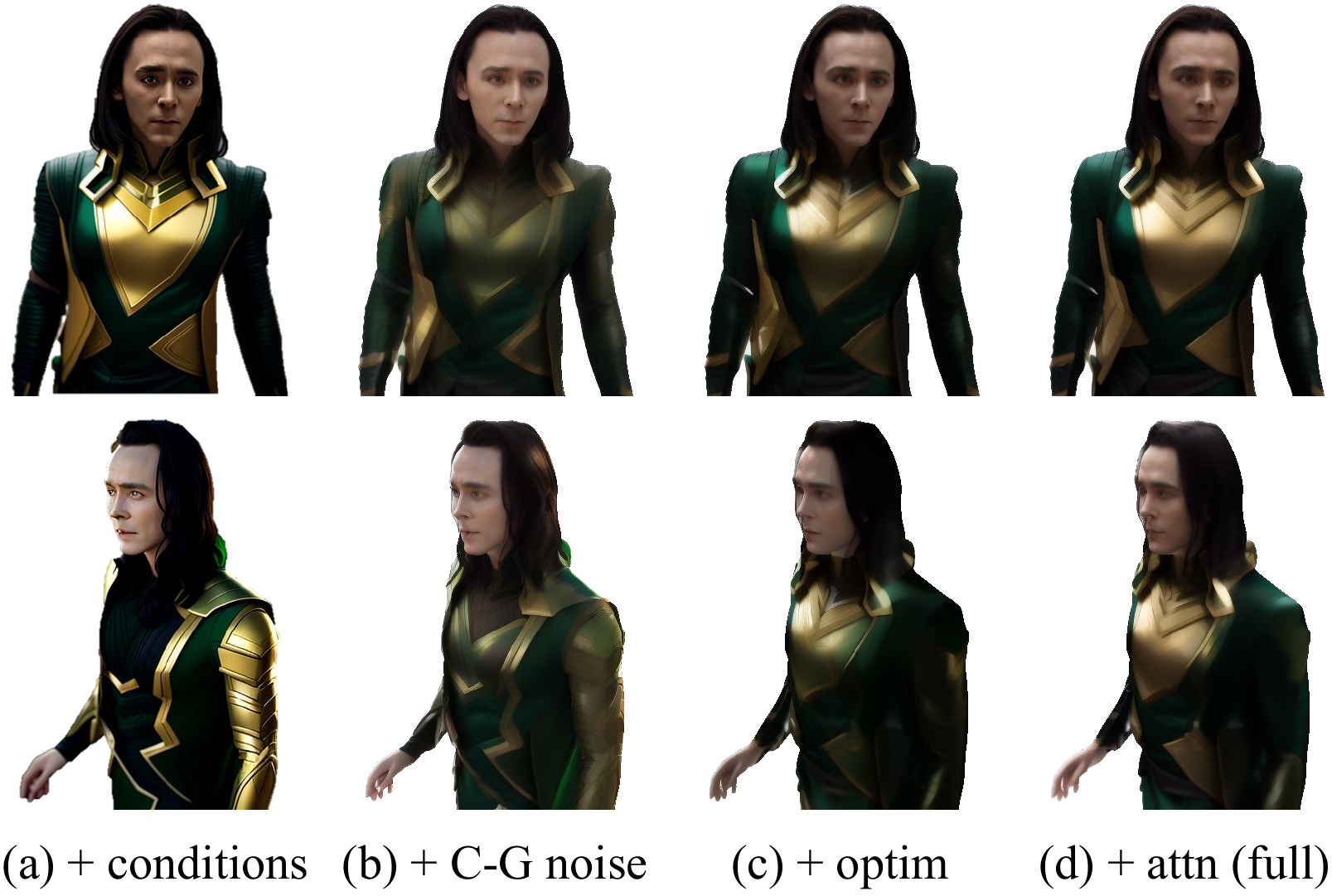} 
	\end{center} 
    \vspace{-20pt}
    \caption{Qualitative evaluation of the multi-view sampling. Note the change in the chest area. (\textit{``Loki, green and gold armor''})}
	\label{fig:fig_qualitative_evaluation_add} 
	\vspace{-10pt}
\end{figure}

\noindent{\textbf{Multi-view Sampling.}}
We evaluate the multi-view sampling process. In Fig.~\ref{fig:fig_qualitative_evaluation_add}, first we condition the SD with depth, normal map and Openpose image of each view, and conduct a vanilla sampling process (\textbf{+ conditions}). The generated results align with the conditions with no view consistency. Then we incorporate the consistency-guided noise (\textbf{+ C-G noise}), and the results show better consistency, yet there remain some blending artifacts in the textures. When we incorporate the optimization of latent codes (\textbf{+ optim}), the blending artifacts are alleviated. Finally, after the self-attention block is modified (\textbf{+ attn}), our full method achieves the best visual effect. We further quantitatively evaluate the view consistency. For each view, we warp the generated image to its neighboring views to compute the PSNR value on the overlapped region. As shown in Tab.~\ref{table:evaluation_consistency}, our full method achieves the best score.

\begin{figure}[t] 
    \vspace{-1ex}
	\begin{center} 
		\includegraphics[width=\linewidth]{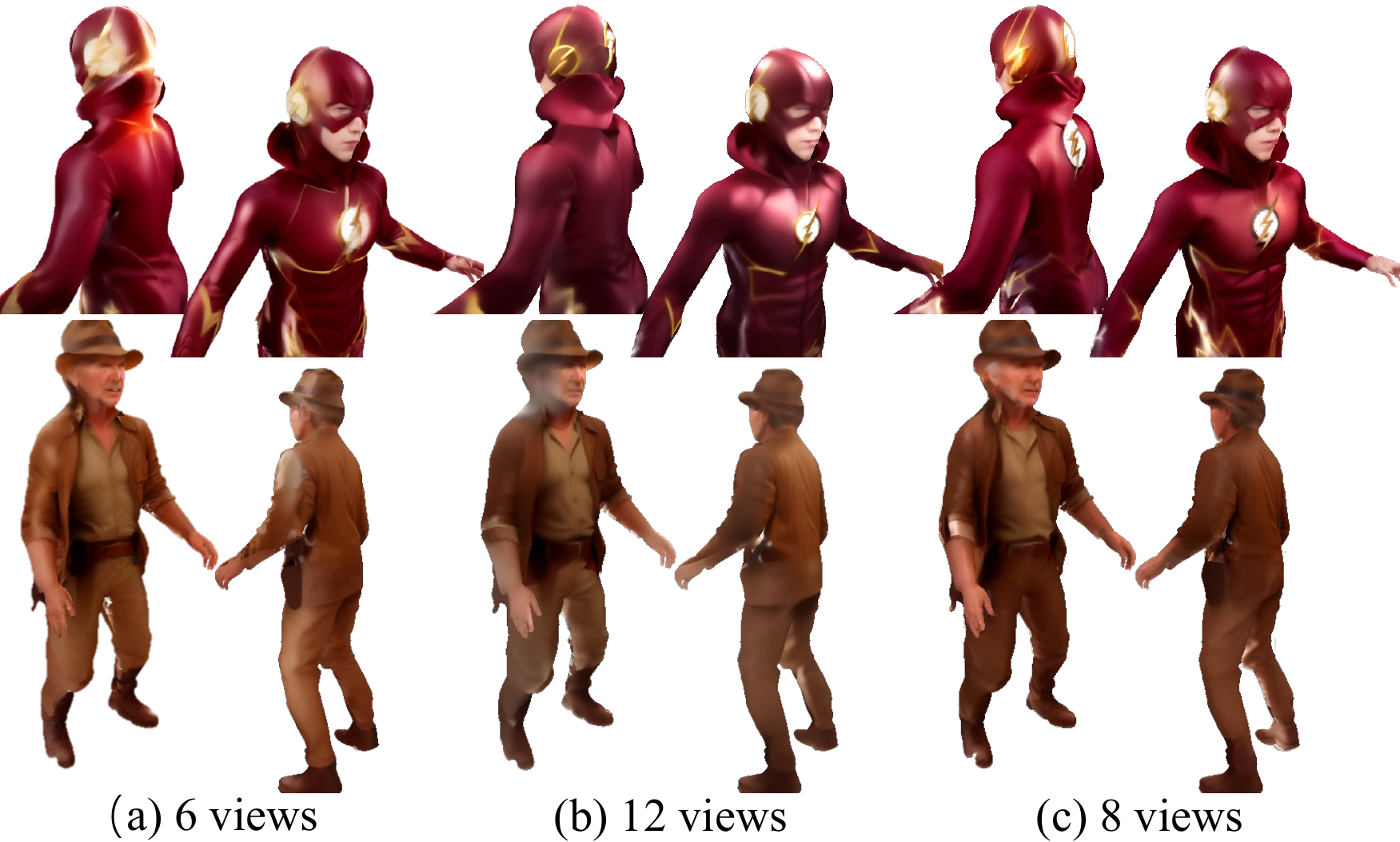} 
	\end{center} 
    \vspace{-20pt}
    \caption{Qualitative evaluation of the number of views on each track. (\textit{``The Flash''} \& \textit{``Harrison Ford''})}
	\label{fig:fig_qualitative_evaluation_cam} 
	\vspace{-10pt}
\end{figure}

\noindent{\textbf{Number of Views.}}
We evaluate the number of views $N$ on each track $R_{fb}, R_{ub}$ qualitatively. As shown in Fig.~\ref{fig:fig_qualitative_evaluation_cam}, when $N=6$, insufficient views can lead to artifacts on NeRF training and final blending results. When $N=12$, the running time can cost more than one hour, and some blurry artifacts may occur due to the imbalanced optimization of latent codes between views. Our setting $N=8$ ensures stable generation while keeping high quality and efficiency.

\begin{figure}[t] 
	\begin{center} 
		\includegraphics[width=\linewidth]{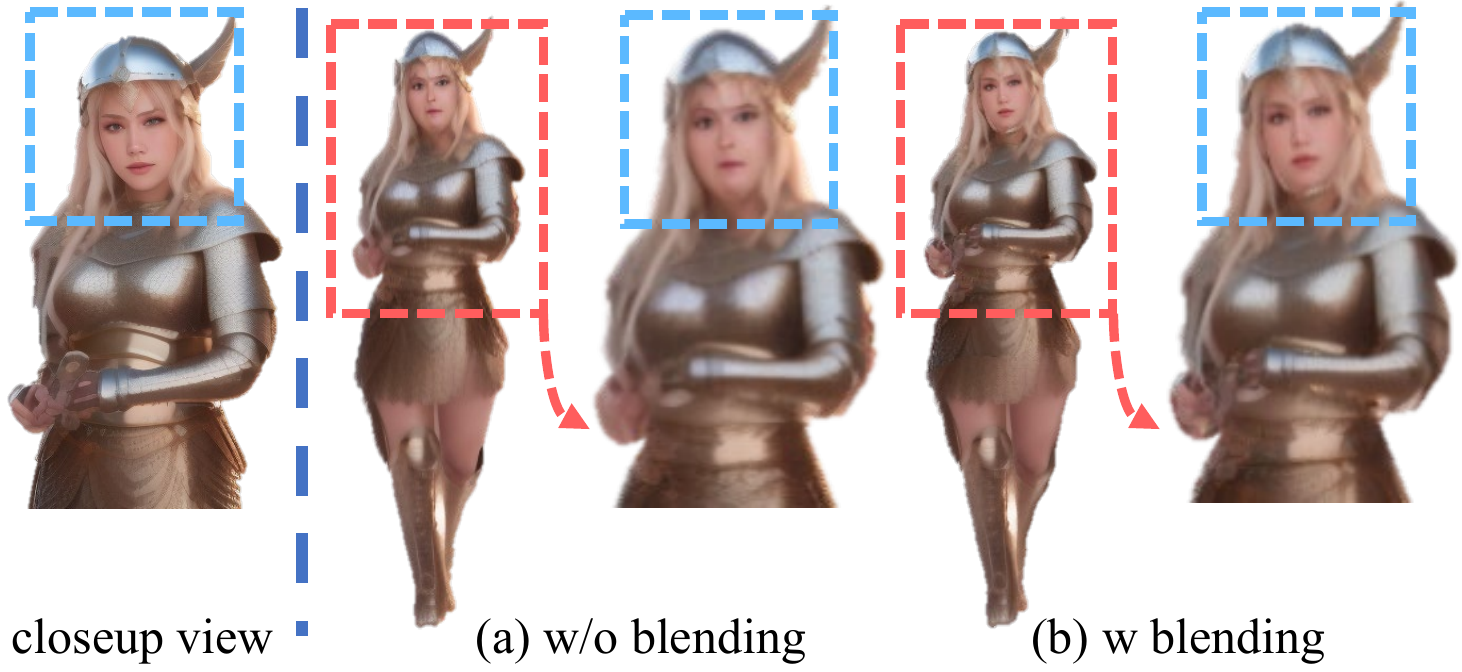} 
	\end{center} 
    \vspace{-20pt}
    \caption{Qualitative evaluation of the neural blending scheme. (\textit{``Valkyrie, in chain mail, skirt armor, and helmet''})}
	\label{fig:fig_qualitative_evaluation_blending} 
	\vspace{-10pt}
\end{figure}

\noindent{\textbf{Blending Scheme.}}
We evaluate our blending scheme qualitatively. As shown in Fig.~\ref{fig:fig_qualitative_evaluation_blending}, the directly generated full-body image (\textbf{w/o blending}) is a little inconsistent with results from the closeup view, especially on the face region. Our blending scheme (\textbf{w blending}) helps improve the generation quality at full-body views. More evaluations are detailed in the supplementation.

\begin{table}[t]
	\begin{center}
		\centering
		\caption{Quantitative evaluation of view consistency.}
		\vspace{-10pt}
		\label{table:evaluation_consistency}
		\resizebox{0.45\textwidth}{!}{
			\begin{tabular}{l|cccc}
				\hline
				Method       
				& \qquad PSNR $\uparrow$ & SSIM $\uparrow$\\
				\hline
				+ conditions \qquad\qquad    &  \qquad 19.297 \qquad &  \qquad  0.9092  \qquad\qquad \\
				+ C-G noise       &  \qquad28.010\qquad    &    \qquad  0.9678   \qquad \qquad  \\
				+ optimization        & \qquad 33.444  \qquad  &    \qquad  0.9851 \qquad \qquad\\
                full        & \qquad34.074  \qquad      &   \qquad  0.9860 \qquad\qquad \\
				\hline
			\end{tabular}
		}
		\vspace{-20pt}
	\end{center}
\end{table}

\subsection{Application} \label{sec:app}
\begin{figure}[t] 
	\begin{center} 
		\includegraphics[width=0.9\linewidth]{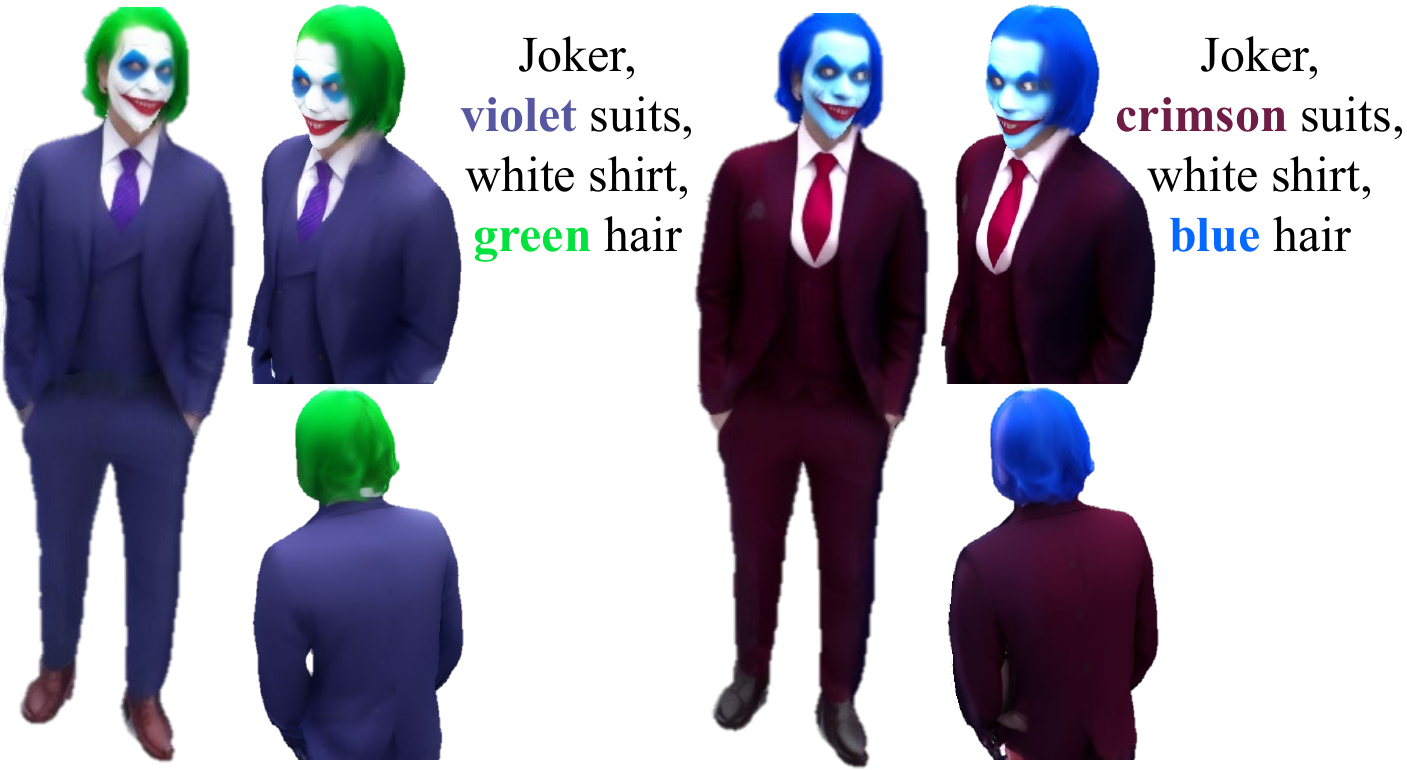} 
	\end{center} 
    \vspace{-20pt}
    \caption{Application: enabling text-guided human asset editing.}
	\label{fig:fig_app_prompt} 
	\vspace{-10pt}
\end{figure} 

\noindent{\textbf{Text-guided Editing.}}
As one of the inherent capabilities of the SD model, text-guided editing can be seamlessly integrated into our MVHuman. As shown in Fig.~\ref{fig:fig_app_prompt}, by modifying the textual description, we can manipulate the color of a character's clothing or hair.

\begin{figure}[t] 
    \vspace{-1ex}
	\begin{center} 
		\includegraphics[width=0.96\linewidth]{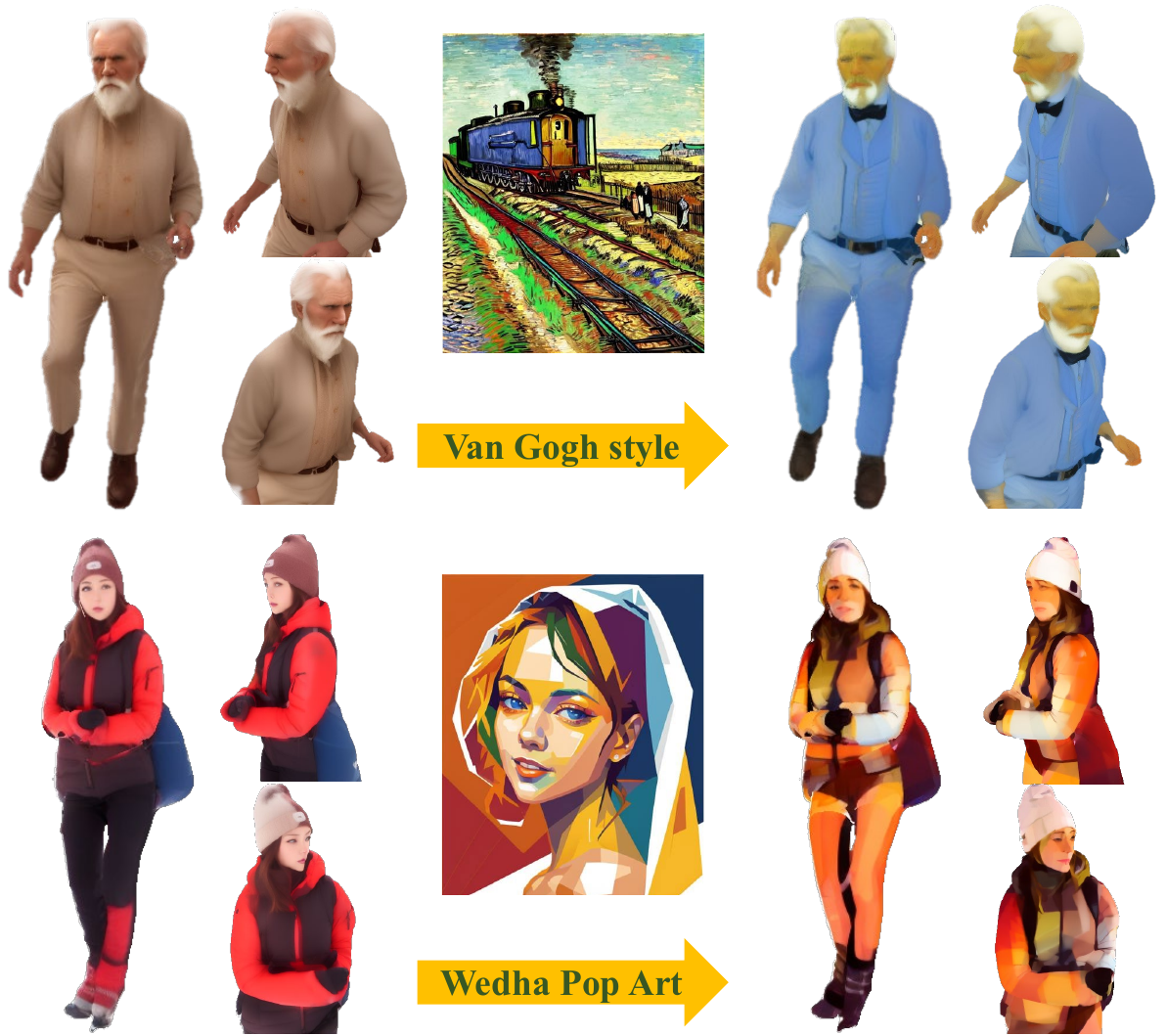} 
	\end{center} 
    \vspace{-20pt}
    \caption{Application: style transfer with LoRA models. (\textit{``Leo Tolstoy, a writer''} \& \textit{``A woman wearing ski clothes''})}
	\label{fig:fig_app_lora} 
	\vspace{-10pt}
\end{figure}

\noindent{\textbf{Style Transfer with LoRA.}}
Our \papername is based on a pre-trained 2D SD model, which leverages an extensive repository of LoRA models. This allows us to seamlessly integrate LoRA models into our method, facilitating tasks such as style transfer. As shown in Fig.~\ref{fig:fig_app_lora}, with two LoRA models loaded to the SD model, the style of the generated results has changed respectively.

\subsection{Discussion}
\noindent{\textbf{Limitations.}} Although MVDream shows promising human generation results, it still has some limitations. First, our method relies on the accuracy of the alignment between the initial coarse mesh and SMPL-X model. It is meaningful to find a way to efficiently obtain these prerequisites. 
Second, textual descriptions often encounter challenges in articulating specific details. This problem may benefit from a combination with image-conditioned diffusion models. Moreover, our method does not support the relighting tasks because it does not take albedo, material into consideration. Thus it is meaningful to have a diffusion model that can generate albedo and various materials given conditions such as semantic segmentation. Also, it could be interesting to combine our method with other 3D representations to better deal with specific generations like hair.

\noindent{\textbf{Social Impact.}} When researching generative technologies, we concern about their potential infringements on intellectual property. There should be heightened legal restrictions on the applications of such technologies. Furthermore, gender and cultural diversity are crucial. It is necessary for any generative technology to ensure inclusivity and avoid stereotypes. In this paper, all our results are carefully selected based on these principles.
\section {Conclusion}
We have presented a novel approach to generate human radiance fields from text prompts, with consistent multi-view images directly sampled from pre-trained 2D diffusion models without any fine-tuning or distilling. The core of our approach is a multi-view sampling process compatible with the 2D diffusion model, which carefully tailors the input conditions, predicted noises and the corresponding latent codes.
It enables consistent multi-view generation by producing the ``consistency-guided noise'' to replace the original predicted noise, explicitly optimizing the latent codes of adjacent views, and modifying the self-attention layer of the pre-trained SD network. Such generated images further benefit producing exquisite human assets with refined geometry and free-view rendering. Our experimental results demonstrate the effectiveness of our approach and we also showcase various 3D applications by seamlessly lifting many editing functions of the original 2D diffusion models to 3D. 
With the above characteristics, we believe that our approach is a critical step towards high-quality 3D human generation, providing a new understanding of the relationship between 2D and 3D generation.

{
    \small
    \bibliographystyle{ieeenat_fullname}
    \bibliography{main}
}

\clearpage
\setcounter{page}{1}
\maketitlesupplementary

\begin{figure*}[t] 
    \vspace{-1ex}
	\begin{center} 
		\includegraphics[width=\linewidth]{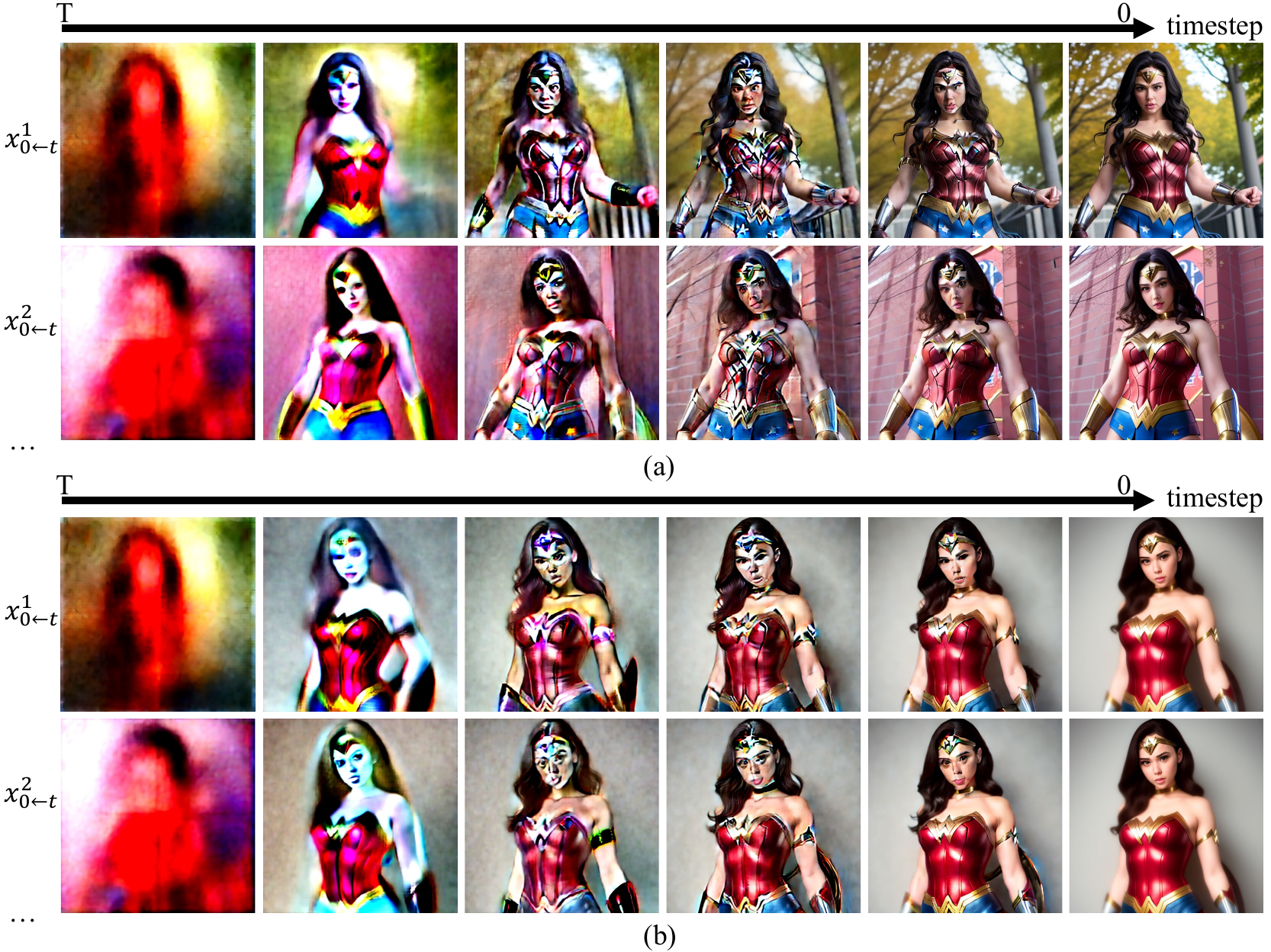} 
	\end{center} 
    \vspace{-20pt}
    \caption{(a) The predicted original signals of simultaneous sampling processes over timesteps. With the vanilla predicted noise for sampling, several simultaneous sampling processes generate various outputs. (b) With the consistency-guided noise for sampling, several simultaneous sampling processes can recover consistent results.}
	\label{fig:2D_sampling} 
	\vspace{-10pt}
\end{figure*} 

In this supplementary material, we first provide additional analysis on the multi-view sampling process in Sec.~\ref{sec:add_analysis}. Then we provide details of our implementation in Sec.~\ref{sec:implementation_details}. More evaluations are discussed in Sec.~\ref{sec:more_comparisons} and Sec.~\ref{sec:more_evaluation}. We further provide more results in Sec.~\ref{sec:more_results}.

\section{Additional Analysis on Multi-view Sampling}
\label{sec:add_analysis}
In this section, we discuss more about the consistency-guided noise of the multi-view sampling process. In sec.~\ref{sec:add_analysis_1}, we showcase a special scenario of multi-view sampling in 2D in which all views are identical. We demonstrate that, with consistency-guided noise, for sampling processes with different initial random noises, they can finally recover consistent results. 
In sec.~\ref{sec:add_analysis_2}, we further analyze how such multi-view sampling process can be transferred into a very different 3D scenario to support generating multi-view consistent images in 3D.

\subsection{Multi-view Sampling in 2D Scenario}\label{sec:add_analysis_1}
In stable diffusion models, for multiple sampling processes with different initial random noises, if we denoise them with their own vanilla predicted noise, the final results will be various because these sampling processes are individual. Yet, we find that we can blend their predicted original signals at sampling timesteps, so as to obtain a consistent prediction and transform it back to the corresponding consistency-guided noises. By replacing the original predicted noises with such consistency-guided ones, various sampling processes can finally recover a consistent signal.

Specifically, assuming we have $N$ identical views, which means there are no warping operations, the scenario would degrade to 2D. At timestep $t$, we have their latent codes: $\{x^k_t| k=1,\ldots,N\}$ and corresponding predicted noises: $\{\epsilon^k_t| k=1,\ldots,N\}$. These noises follow a normal distribution of $\mathcal{N}(\textbf{0}, \textbf{I})$. With following equation: 
\begin{equation}
{x}^k_{0\leftarrow t}=\frac{x^k_t-\sqrt{1-\alpha_t} \epsilon^k_t}{\sqrt{\alpha_t}},
\label{eqn:eqn_noise2origin_sup} 
\end{equation}
we can obtain corresponding predicted original signals at step t: $\{x^k_{0\leftarrow t}| k=1,\ldots,N\}$. For each $x^k_{0\leftarrow t}$, it should follow the distribution of $\mathcal{N}(\frac{x^k_t}{\sqrt{\alpha_t}}, \frac{1-\alpha_t}{\alpha_t}\textbf{I})$ respectively. Note that these predicted original signals have various means but share the same variance. Since there is no warping operation, we can directly calculate the average result $\bar{{x}}_{0\leftarrow t}$ following:
\begin{equation}
    \bar{{x}}_{0\leftarrow t}=\frac{1}{N} (x^1_{0\leftarrow t} + \ldots + x^N_{0\leftarrow t})
    \label{eqn:average} 
\end{equation}
which should follow the following distribution:
\begin{equation}
\begin{aligned}
    \bar{{x}}_{0\leftarrow t} &\sim \mathcal{N}(\frac{1}{N\sqrt{\alpha_t}} (x^1_t + \ldots + x^N_t), \frac{1\times N}{N^2} \frac{1-\alpha_t}{\alpha_t}\textbf{I}) \\
    &\sim \mathcal{N}(\frac{x^1_t + \ldots + x^N_t}{N\sqrt{\alpha_t}}, \frac{1}{N} \frac{1-\alpha_t}{\alpha_t}\textbf{I}).
\end{aligned}
\end{equation}
 In order to maintain sampling quality, we need to maintain the variance of the distribution unchanged as $\frac{1-\alpha_t}{\alpha_t}\textbf{I}$. Note that for each $x^k_{0\leftarrow t}$, its mean is known and related to $x^k_t$. Thus we can easily scale its noise part by $\sqrt{N}$ so that each new $x^k_{0\leftarrow t}$ follows a distribution of $\mathcal{N}(\frac{x^k_t}{\sqrt{\alpha_t}}, N\frac{1-\alpha_t}{\alpha_t}\textbf{I})$. Then with Eqn.~\ref{eqn:average}, we can get a new average result $\Tilde{{x}}_{0\leftarrow t}$:
\begin{equation}
\begin{aligned}
    \Tilde{{x}}_{0\leftarrow t} &\sim \mathcal{N}(\frac{1}{N\sqrt{\alpha_t}} (x^1_t + \ldots + x^N_t), \frac{1\times N}{N^2} N\frac{1-\alpha_t}{\alpha_t}\textbf{I}) \\
    &\sim \mathcal{N}(\frac{x^1_t + \ldots + x^N_t}{N\sqrt{\alpha_t}}, \frac{1-\alpha_t}{\alpha_t}\textbf{I}).
\end{aligned}
\end{equation}
With $\Tilde{{x}}_{0\leftarrow t}$ and $x^k_t$, we can then obtain $N$ new consistency-guided noises with Eqn.~\ref{eqn:eqn_noise2origin_sup}, denoted as $\epsilon^{' k}_t$, $k=1,\ldots,N$: 
\begin{equation}
\begin{aligned}
    \epsilon^{' k}_t &= \frac{x^k_t - \sqrt{\alpha_t}\Tilde{{x}}_{0\leftarrow t}}{\sqrt{1-\alpha_t}} \\
    &\sim \mathcal{N}(\frac{1}{\sqrt{1-\alpha_t}} (x^k_t - \frac{x^1_t + \ldots + x^N_t}{N}), \textbf{I}).
\end{aligned}
\end{equation}
We can observe that the mean of each consistency-guided noise $\epsilon^{' k}_t$ is not $\textbf{0}$ and thus provides a direction towards a common target for each sampling process. 

As shown in Fig.~\ref{fig:2D_sampling}(a), with vanilla predicted noise for sampling, several simultaneous sampling processes with different initial noise will generate various individual results. In Fig.~\ref{fig:2D_sampling}(b), with the consistency-guided noise for sampling, these sampling processes can finally recover high-quality consistent results.

\subsection{Multi-view Sampling in 3D Scenario}\label{sec:add_analysis_2}
With the multi-view sampling in 2D, we then aim to lift it into the 3D scenario so as to generate multi-view consistent images, which can further help reconstruct 3D results. In 3D scenario, the calculation of consistency-guided noise differs in three aspects:
%

1) The overlapped region differs between each pair of views. Thus, for each view as a target view, it only blends its predicted original signal with that from neighboring views that have common overlapped regions to obtain a consistent prediction.

2) In 3D, we cannot directly warp latent codes and predicted original signals between views because view consistency in image space is not equivalent to that in latent space. Also, it is $64\times 64$ low resolution in latent space, and warping operations in such low resolution can lead to misalignment. In order to obtain view consistency in image space, we transform the predicted original signals from source views $v_j,j=j_1,j_2,\ldots$ to target view $v_i$ with the following decode-warp-encode operation:
\begin{equation}
    {x}^{' j}_{0\leftarrow t} = \mathcal{E}(W^i_{j}(\mathcal{D}({x}^{j}_{0\leftarrow t}))).
    \label{eqn:eqn_space_transfer_sup} 
\end{equation}

3) Notably, the decode-warp-encode operation will somehow change the distributions of the predicted original signals $x^j_{0\leftarrow t}$, making it not easy to measure the means of $x^{' j}_{0\leftarrow t}$. Since $x^j_{0\leftarrow t}$ have the same variance $\frac{1-\alpha_t}{\alpha_t}\textbf{I}$, we can still assume $x^{' j}_{0\leftarrow t}$ follow normal distributions of the same variance $\sigma^2$ because the warping operation will not modify pixel value. Based on such assumption, we design an empirical formula to enhance the variance of the distribution of the weighted average which is then used to calculate consistency-guided noise.

%

\begin{figure*}[t] 
	\begin{center} 
		\includegraphics[width=\linewidth]{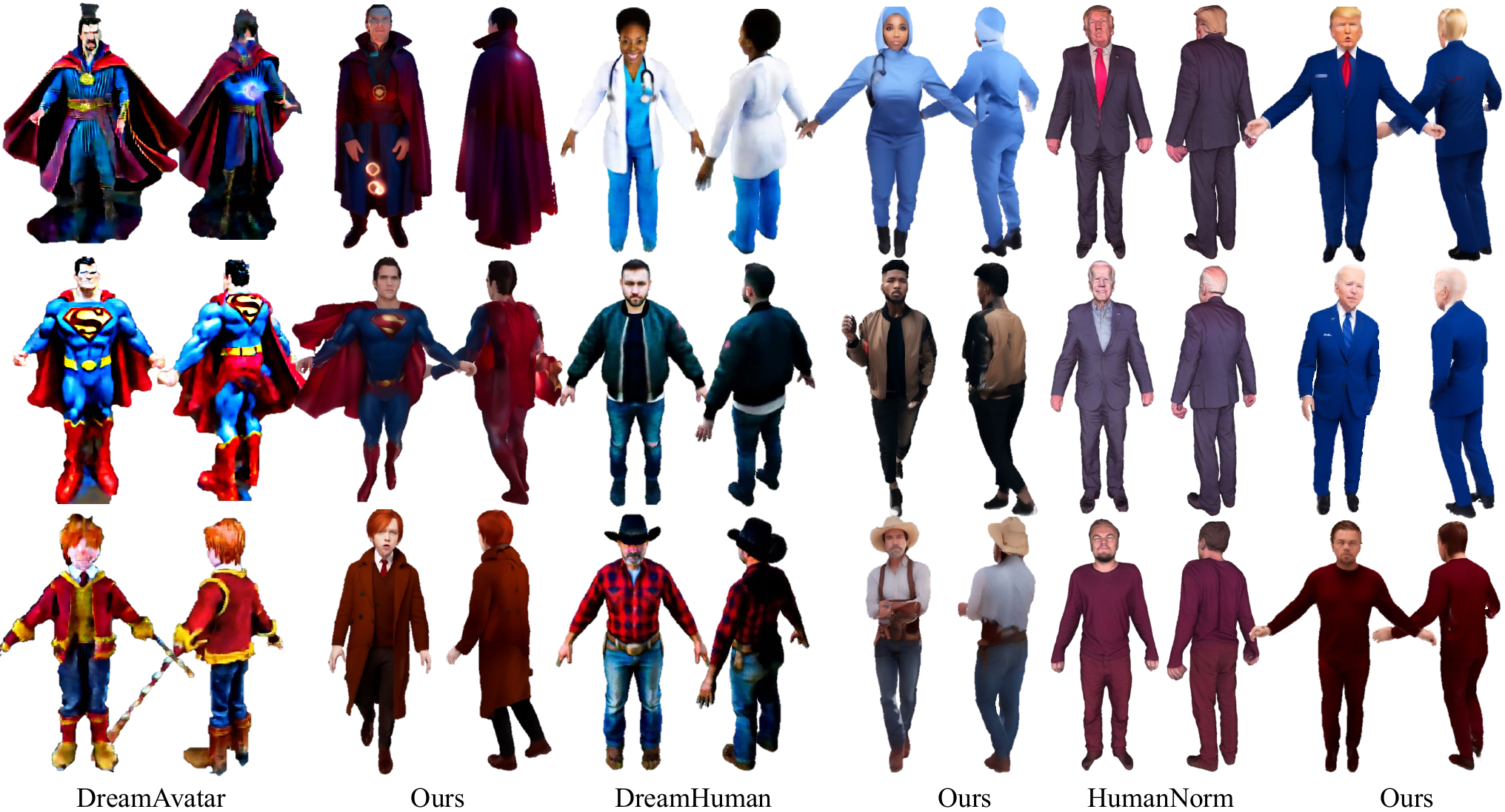} 
	\end{center} 
    \vspace{-20pt}
    \caption{Qualitative comparison with DreamAvatar, DreamHuman, HumanNorm. Our results demonstrate photo-realistic appearance while avoiding oversaturation and Janus failure. The prompts of the first column from top to down are \textit{``Doctor Strange''}(* case using scan mesh), \textit{``Superman''}, \textit{``Ronald Weasley''}. The prompts of the second column from top to down are \textit{``a black female surgeon''}, \textit{``a man wearing a bomber jacket''}, \textit{``a cowboy''}. The prompts of the third column from top to down are \textit{``Donald Trump''}, \textit{``Joe Biden''}, \textit{``Leonardo DiCaprio in a maroon long sleeve top''}.}
	\label{fig:comparison_more} 
\end{figure*} 

\section{Implementation Details}
\label{sec:implementation_details}
We test our MVHuman using NVIDIA RTX 3090 GPUs. The ControlNet models we use are the official version of v1.1. The stable diffusion models used are the official version of v1.5 and the open-source ChilloutMix model which is a fine-tuned SD v1.5 model with a better ability to generate human images. The sampler tested are DDIM~\cite{song2020denoising} and Dpm-solver++~\cite{lu2022dpm}, and the number of sampling timesteps is set as 150. For all the results, we generate images of $512\times 512$ resolution. The parameters in Eqn.~\ref{eqn:eqn_transfer} are $w_s=0.2$, $w_c=1.0$. Upon completion of $0\%-20\%$ of the multi-view sampling process, we initiate the adoption of consistency-guided noise for sampling at timesteps (abbreviated as a C-G step), but the original predicted noise is still adopted one step every one C-G step to improve quality. For the latent codes optimization, we perform it every 4 timesteps in our method. Our full method takes about 46 minutes to generate a case.


%

\section{More Comparisons}
\label{sec:more_comparisons}
In this section, we provide more comparison results with SOTA prompt-guided human generation methods (\ref{sec:more_comparisons_human_generation}), single-image human reconstruction methods (\ref{sec:more_comparisons_reconstruction}), and recent multi-view diffusion model (\ref{sec:more_comparisons_mvd}).

\subsection{Comparison with Human Generation Methods}
\label{sec:more_comparisons_human_generation}
We further compare our method with more state-of-the-art text-guided 3D human generation methods, DreamAvatar~\cite{cao2023dreamavatar}, DreamHuman~\cite{kolotouros2023dreamhuman}, HumanNorm~\cite{huang2023humannorm}. These methods are not open-source currently, so the compared cases are selected from their galleries. As illustrated in Fig.~\ref{fig:comparison_more}, DreamAvatar suffers from oversaturation and Janus problem. DreamHuman demonstrates good color saturation, however, its low resolution imposes constraints on the final quality. HumanNorm produces mesh with texture, but its results lack photo-realistic fidelity.

\begin{figure}[t] 
    \vspace{-1ex}
	\begin{center} 
		\includegraphics[width=1.\linewidth]{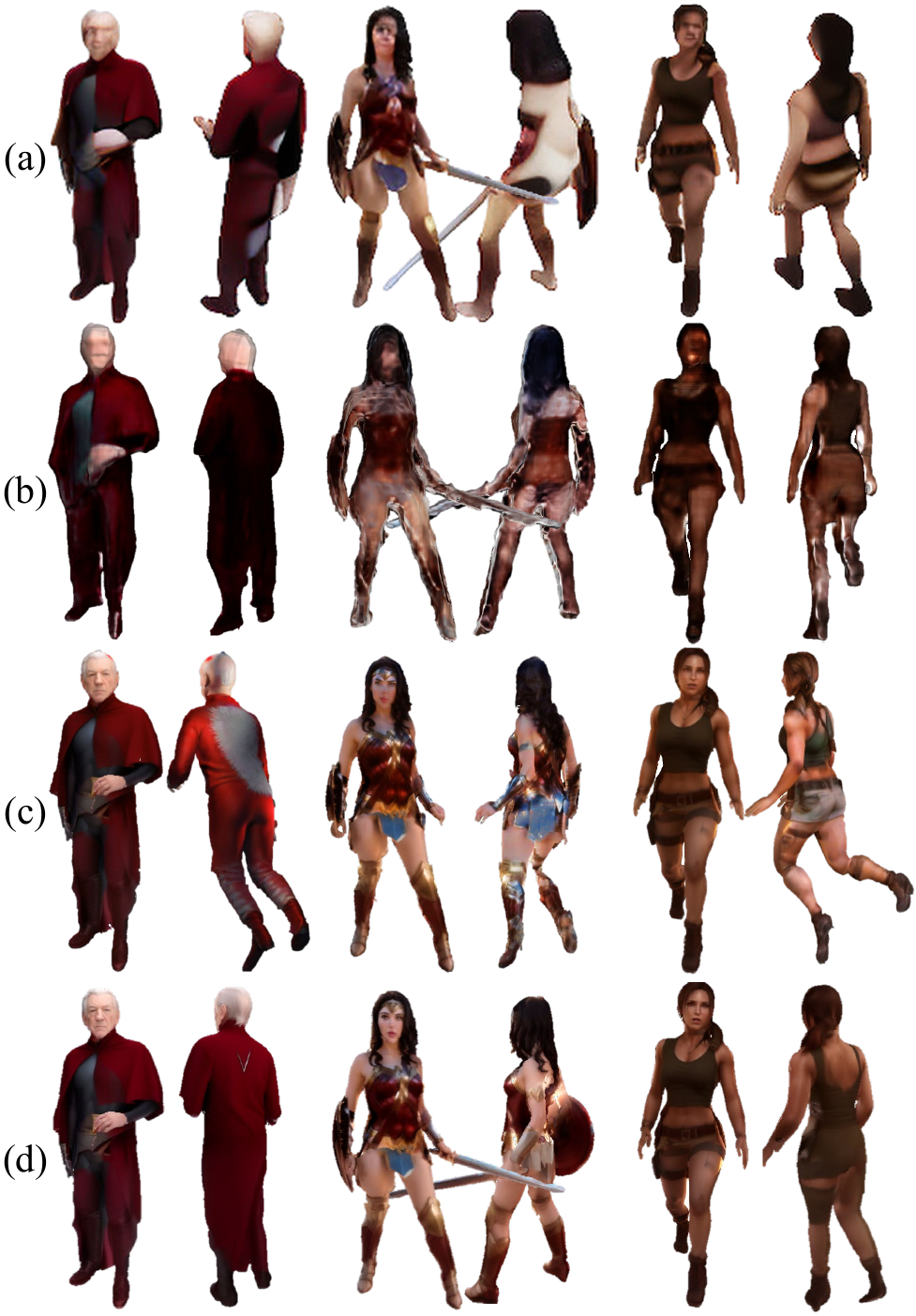} 
	\end{center} 
    \vspace{-20pt}
    \caption{Qualitative comparison with image-to-3D methods. (a) SyncDreamer; (b) Wonder3D; (c) TeCH; (d) Ours.}
	\label{fig:fig_comparison_imageto3D} 
	\vspace{-10pt}
\end{figure} 

 \begin{figure}[t] 
    \vspace{-1ex}
	\begin{center} 
		\includegraphics[width=\linewidth]{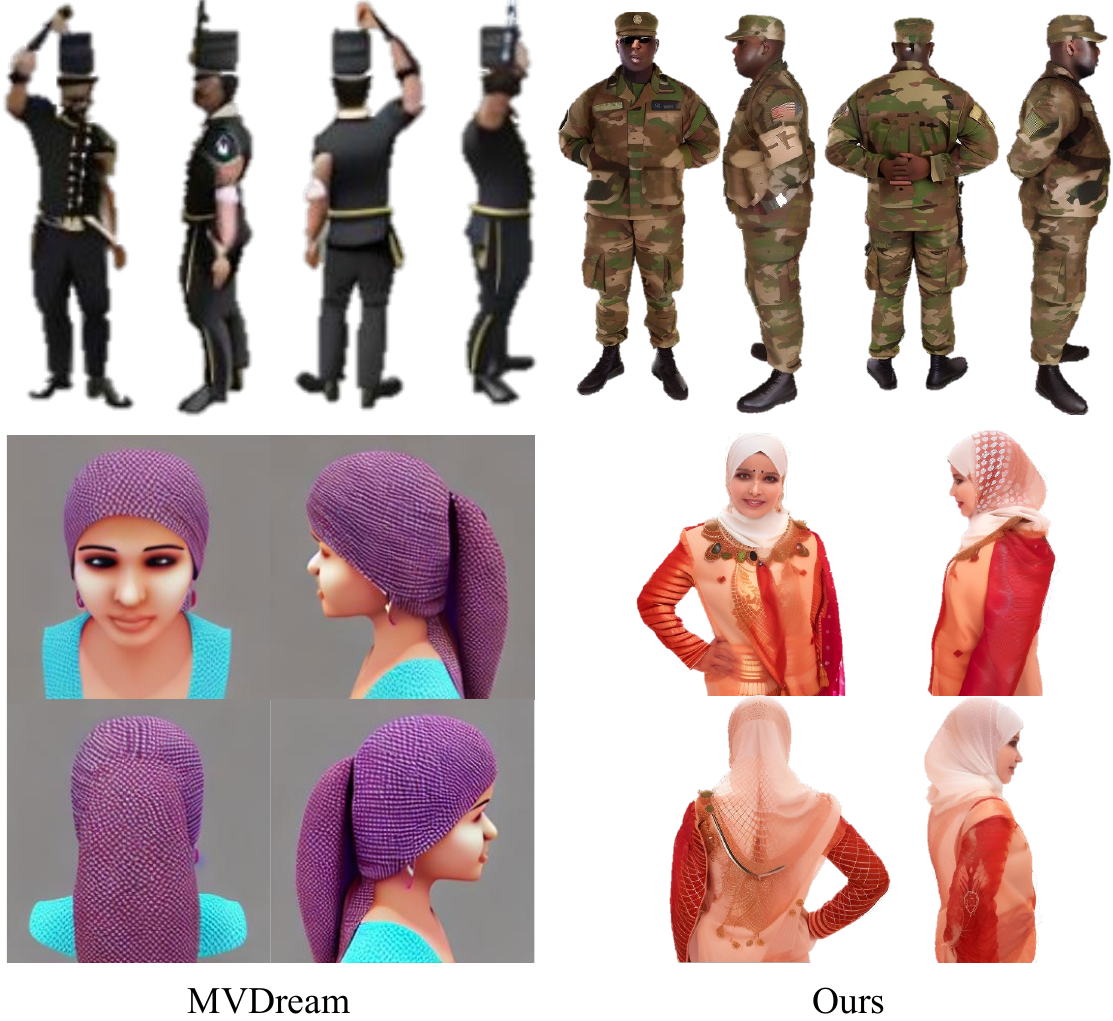} 
	\end{center} 
    \vspace{-20pt}
    \caption{Qualitative comparison with MVDream. Our appearance is more photo-realistic. The prompts from top to down are \textit{``a black army soldier''}, \textit{``an Indian woman wearing a hood, upper body''}. (* Both of ours are cases using scan meshes)}
	\label{fig:fig_comparison_mvd} 
	\vspace{-10pt}
\end{figure}

\subsection{Comparison with Image-to-3D Methods}
\label{sec:more_comparisons_reconstruction}
We further qualitatively compare our method with state-of-the-art image-to-3D methods. As illustrated in Fig.~\ref{fig:fig_comparison_imageto3D}, SyncDreamer~\cite{liu2023syncdreamer} and Wonder3D~\cite{long2023wonder3d} both take an image as input and output multi-view images for reconstruction. They can generate human body images from various views, but the appearance has blurry artifacts, particularly in the facial part. TeCH~\cite{huang2023tech} takes more than 3 hours on its network fine-tuning and SDS-based optimization. It faithfully recovers the input image on frontal view, but exhibits some inconsistencies on other views. Our method achieves high-quality generation while maintaining high view consistency.

\subsection{Comparison with Multi-view Diffusion Model}
\label{sec:more_comparisons_mvd}
We further compare our method with the direct outcomes of MVDream. Given a text prompt, MVDream can generate four images of $256\times 256$ resolution in less than 1 minute. MVHuman can also be easily customized to generate specific four views. As illustrated in Fig.~\ref{fig:fig_comparison_mvd}, our method can generate high-quality results while the results of MVDream tend to be cartoon-like. Notably, our method takes less than 7 minutes for a case under the situation without any code optimization.

\section{More Ablation Study}
\label{sec:more_evaluation}

\begin{figure}[t] 
    \vspace{-1ex}
	\begin{center} 
		\includegraphics[width=\linewidth]{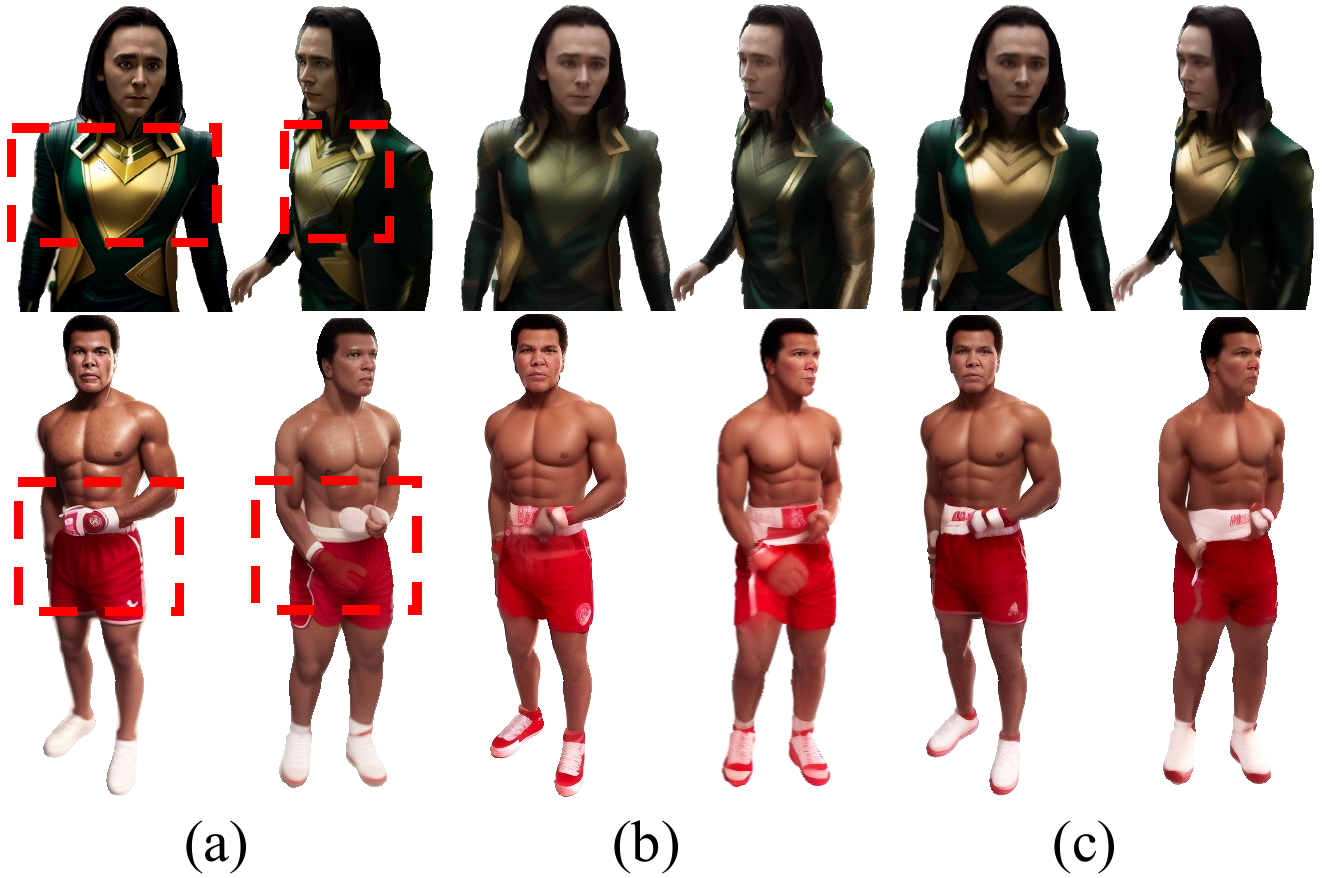} 
	\end{center} 
    \vspace{-20pt}
    \caption{Qualitative evaluation of the multi-view sampling. (a) w/o C-G noise; (b) w/o optimization; (c) full}
	\label{fig:fig_more_ablation} 
	\vspace{-10pt}
\end{figure} 

\begin{table}[t]
	\begin{center}
		\centering
		\caption{Quantitative evaluation of view consistency.}
		\vspace{-10pt}
		\label{table:evaluation_more_ablation}
		\resizebox{0.45\textwidth}{!}{
			\begin{tabular}{l|cccc}
                \hline
                Method       
				& \qquad PSNR $\uparrow$ & SSIM $\uparrow$\\
				\hline
                w/o C-G noise         &   \qquad   29.497   \qquad   &   \qquad    0.9630   \qquad\qquad    \\
                w/o optimization       & \qquad     28.121    \qquad  &    \qquad   0.9699   \qquad\qquad  \\
                full        & \qquad34.074  \qquad      &   \qquad  0.9860 \qquad\qquad \\
				\hline
			\end{tabular}
		}
		\vspace{-20pt}
	\end{center}
\end{table}

\begin{figure}[t] 
    \vspace{-1ex}
	\begin{center} 
		\includegraphics[width=\linewidth]{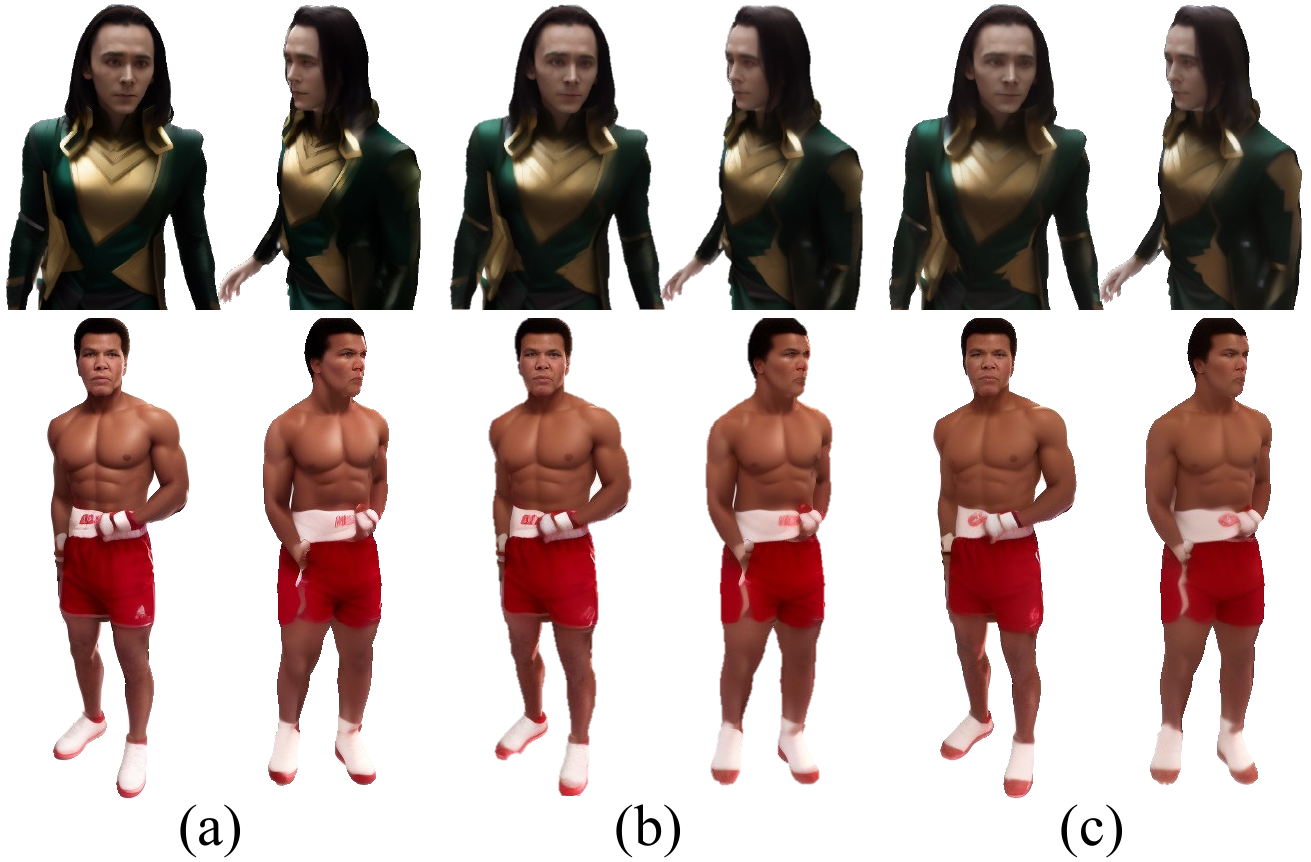} 
	\end{center} 
    \vspace{-20pt}
    \caption{Qualitative evaluation of mixed sampling steps. (a) one original sampling step every one C-G step; (b) one original sampling step every two C-G steps; (c) all with C-G steps.}
	\label{fig:fig_more_ablation_noise} 
	\vspace{-10pt}
\end{figure}

\begin{figure}[t] 
    \vspace{-1ex}
	\begin{center} 
		\includegraphics[width=\linewidth]{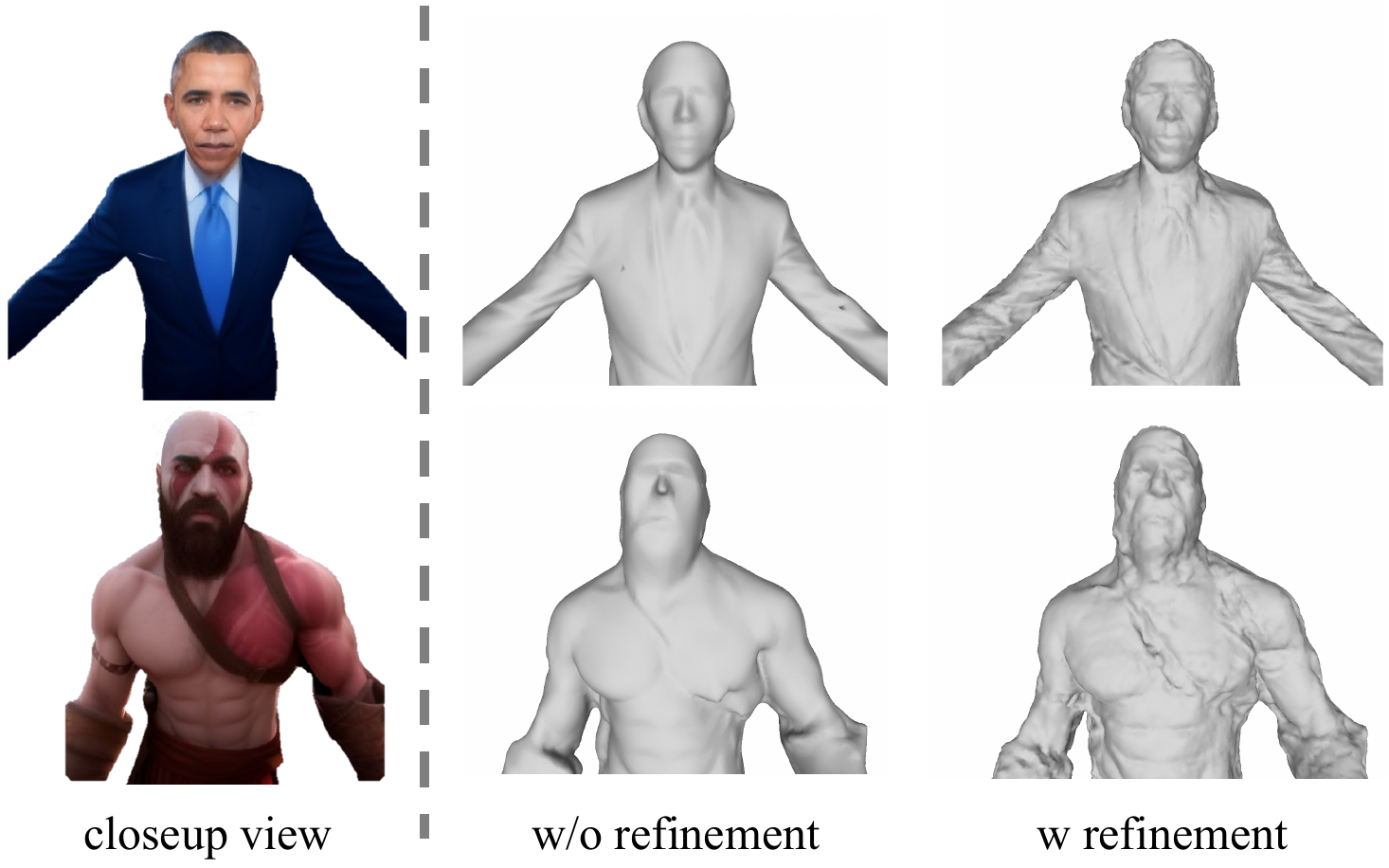} 
	\end{center} 
    \vspace{-20pt}
    \caption{Qualitative evaluation of geometry refinement. With geometry refinement, we can obtain fine-grained geometry having details aligned with the appearance in the images.}
	\label{fig:fig_qualitative_evaluation_geometry} 
	\vspace{-10pt}
\end{figure}

\begin{figure*}[t] 
	\begin{center} 
		\includegraphics[width=\linewidth]{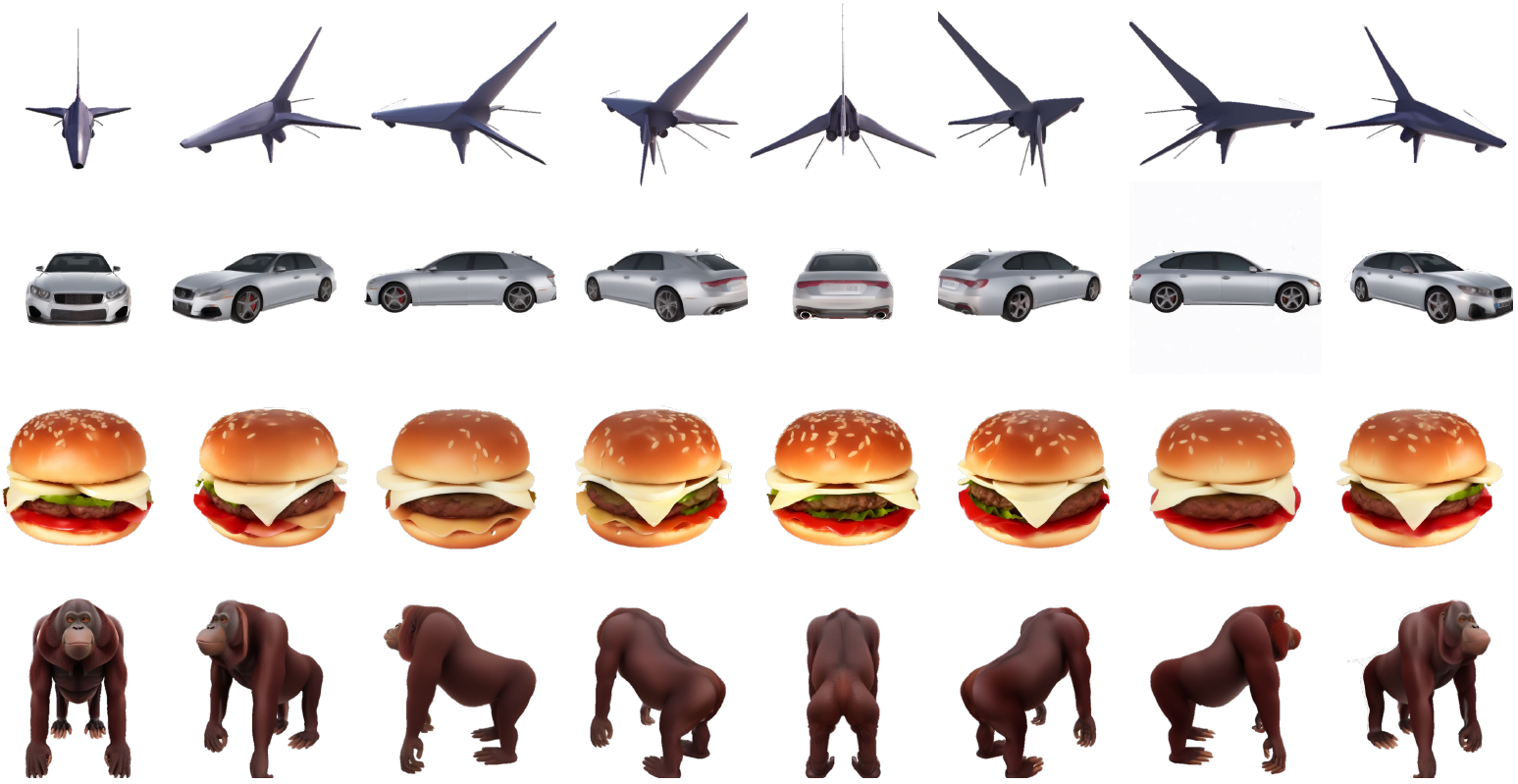} 
	\end{center} 
    \caption{More results on general objects. The prompts from top to down are \textit{``a starship''}, \textit{``a silver car''}, \textit{``a delicious hamburger''}, \textit{``an orangutan''}.}
	\label{fig:more_results} 
\end{figure*} 

\subsection{Multi-view Sampling}
We further evaluate the multi-view sampling process. In Fig.~\ref{fig:fig_more_ablation}, without the consistency-guided noise (\textbf{w/o C-G noise}), though the outcomes appear satisfactory from each view, the overall consistency is lacking because the optimization alone can not provide enough constraints on view consistency.  Without the latent codes optimization (\textbf{w/o optimization}), since we mix the use of the original predicted noise and the C-G noise, the outcomes are also not strictly view-consistent. Our full method achieves high view consistency and maintains high quality at the same time. As shown in Tab.~\ref{table:evaluation_more_ablation}, the full method achieves the best score.

We then evaluate the ratio between sampling steps using the original predicted noise (original sampling step) and steps using the C-G noise (C-G step). Our full sampling process adopts one original sampling step every one C-G step (a step using C-G noise for sampling). As shown in Fig.~\ref{fig:fig_more_ablation_noise}(a), with this setting, we can obtain results that are both high-quality and view-consistent. In Fig.~\ref{fig:fig_more_ablation_noise}(b), the sampling process adopts one original sampling step every two C-G steps, and in Fig.~\ref{fig:fig_more_ablation_noise}(c), all sampling steps are C-G steps. We can observe that as the proportion of C-G steps increases, it sacrifices some visual effects in exchange for higher view consistency.

\subsection{Geometry Refinement}
We additionally evaluate the geometry refinement of the post-process. As illustrated in Fig.~\ref{fig:fig_qualitative_evaluation_geometry}. The initial geometry appears to be smooth. The geometry refinement adds fine-grained details to the initial geometry. Meanwhile, these details also align well with the appearance depicted in the images.


\section{More Results}
\label{sec:more_results}

We also test our method on generating general objects by customizing the multi-view sampling process to generate images of 8 views in a single circular track. The conditions for ControlNet are depth and normal maps. As illustrated in Fig.~\ref{fig:more_results}, our method can also be applied to generate general objects. However, it may fail when the given conditions are ambiguous (e.g. the provided geometry is too smooth).

\section{Ethics Statement}
\label{sec:Ethics_Statement}
Though human eyes can distinguish real-world photos and results of MVHuman, we worry about the potential ethical risk. Thus we want to state again that MVHuman should not be utilized to deceive those who are unfamiliar with this domain. Moreover, like all other generative methods, MVHuman should also not be utilized to generate results that violate the diversity of gender, race, and culture.

\end{document}